\documentclass[10pt,twocolumn,letterpaper]{article}

\usepackage{cvpr}
\usepackage{times}
\usepackage{epsfig}
\usepackage{graphicx}
\usepackage{amsmath}
\usepackage{amssymb}

\usepackage{caption}
\usepackage{subcaption}
\usepackage{epsfig}
\usepackage{amsmath}
\usepackage{amssymb}
\usepackage{booktabs}
\usepackage{rotating}
\usepackage{algorithm} 
\usepackage{algpseudocode}
\usepackage{multirow}

\usepackage{enumitem}
\usepackage{enumerate}

\newif\ifshowcomments
\showcommentstrue

\ifshowcomments
\newcommand{\TODO}[1]{{\color{red}{[TODO: #1]}}}
\newcommand{\revised}[1]{{\color[rgb]{0.2,0.7,0.2}{#1}}}
\newcommand{\lzhu}[1]{{\color[rgb]{0.7,0.7,0}{[ZL: #1]}}}
\newcommand{\phil}[1]{{\color[rgb]{0.9,0.1,0.1}{[PF: #1]}}}
\newcommand{\michael}[1]{{\color[rgb]{0.6,0.9,0.1}{[MSB: #1]}}}

\else
\newcommand{\TODO}[1]{}
\newcommand{\revised}[1]{}
\newcommand{\lzhu}[1]{}
\newcommand{\phil}[1]{}
\newcommand{\michael}[1]{}
\fi

\usepackage[pagebackref=true,breaklinks=true,letterpaper=true,colorlinks,bookmarks=false]{hyperref}

\cvprfinalcopy 

\def\cvprPaperID{2194} 

\ifcvprfinal\fi
\begin{document}

\title{Direction-aware Spatial Context Features for Shadow Detection}
\author{Xiaowei Hu$^{1, \ast}$, Lei Zhu$^{2,}$\thanks{Joint first authors}\ , Chi-Wing Fu$^{1,3}$, Jing Qin$^{2}$, and Pheng-Ann Heng$^{1, 3}$\\
	$^1$ Department of Computer Science and Engineering, The Chinese University of Hong Kong
	\\$^2$ Centre for Smart Health, School of Nursing, The Hong Kong Polytechnic University
	\\$^3$ Guangdong Provincial Key Laboratory of Computer Vision and Virtual Reality Technology,
	\\Shenzhen Institutes of Advanced Technology, Chinese Academy of Sciences, China}

\maketitle
\thispagestyle{empty}


\begin{abstract}
Shadow detection is a fundamental and challenging task, since it requires an understanding of global image semantics and there are various backgrounds around shadows.
This paper presents a novel network for shadow detection by analyzing image context in a direction-aware manner.
To achieve this, we first formulate the direction-aware attention mechanism in a spatial recurrent neural network (RNN) by introducing attention weights when aggregating spatial context features in the RNN.
By learning these weights through training, we can recover direction-aware spatial context (DSC) for detecting shadows.
This design is developed into the DSC module and embedded in a CNN to learn DSC features at different levels.
Moreover, a weighted cross entropy loss is designed to make the training more effective.
We employ two common shadow detection benchmark datasets and perform various experiments to evaluate our network.
Experimental results show that our network outperforms state-of-the-art methods and achieves 97\% accuracy and 38\% reduction on balance error rate.
\end{abstract}


\section{Introduction}
\label{sec::introduction}

Shadow is a monocular cue in human vision for depth and geometry perception.
Knowing where the shadow is, on the one hand, allows us to acquire the lighting direction~\cite{lalonde2009estimating} and scene geometry~\cite{okabe2009attached,karsch2011rendering}, as well as the camera location and parameters~\cite{junejo2008estimating}.
However, the presence of shadow, on the other hand, could deteriorate the performance of many fundamental computer vision tasks, such as object detection and tracking~\cite{cucchiara2003detecting,nadimi2004physical}.
Hence, shadow detection has long been a fundamental problem in computer vision.


Existing methods detect shadows by developing physical models of color and illumination~\cite{finlayson2006removal,finlayson2009entropy}, or by using data-driven approaches based on hand-crafted features~\cite{huang2011characterizes,lalonde2010detecting,zhu2010learning} or learned features~\cite{khan2014automatic,vicente2016large,nguyen2017shadow}.
While state-of-the-art methods have already achieved accuracy of 87\% to 90\% on two benchmark datasets~\cite{vicente2016large,zhu2010learning}, they may misrecognize black objects as shadows and miss unobvious shadows.
These situations are revealed by the balance error rate (BER), which equally considers shadow and non-shadow regions; see Section~\ref{sec:experiments} for quantitative comparison results.
%


\begin{figure}[!t]
	\centering
	\includegraphics[width=0.8\linewidth]{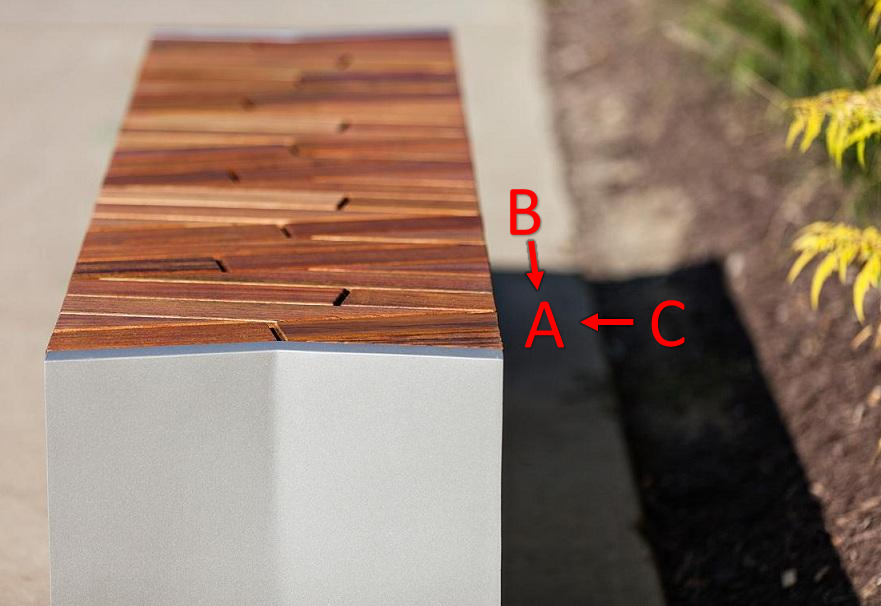}
	\caption{In this example image, region B would give a stronger indication that A is a shadow compared to region C.
	This motivates us to analyze the global image context in a direction-aware manner for shadow detection.}
	\label{fig:motivation}	
\end{figure}

Shadow detection requires an understanding of global image semantics, as shown very recently in~\cite{nguyen2017shadow}.
To improve the understanding, we propose to analyze the image (or spatial) context in a {\em direction-aware\/} manner.
Taking region A in Figure~\ref{fig:motivation} as an example, comparing it with regions B and C, region B would give a stronger indication that A is a shadow as compared to region C.
Hence, spatial contexts along different directions would give different amount of contributions in suggesting the presence of shadows.

To take directional variance into account when reasoning the spatial contexts, we first design a network module called the {\em direction-aware spatial context} (DSC) module, or {\em DSC module\/} for short, by adopting a spatial recurrent neural network (RNN) to aggregate spatial contexts in four principal directions, and formulating the direction-aware attention mechanism in the RNN to learn attention weights for each direction.
Then, we embed multiple copies of this DSC module in a convolutional neural network to learn DSC features in different layers (scales).
After that, we combine the DSC features with convolutional features to predict a score map for each layer, and fuse the score maps into the final shadow detection map.
The whole network is trained in an end-to-end manner with a weighted cross entropy loss.
We summarize the major contributions of this work below:
\begin{itemize}[]

\vspace*{-0.25mm}
\item
First, we design a novel attention mechanism in a spatial RNN and construct the DSC module to learn spatial contexts in a direction-aware manner.

\item
Second, we present a new shadow detection network that adopts multiple DSC modules to learn direction-aware spatial contexts in different layers.
A weighted cross entropy loss is designed to balance the detection accuracy in shadow and non-shadow regions.
This network has potential for use in other applications such as saliency detection and semantic segmentation.

\item
Third, we evaluate our network on two benchmark sets and compare it with several state-of-the-art methods on shadow detection, saliency detection, and semantic image segmentation.
Results show that our network outperforms previous methods with over 97\% accuracy and 38\% reduction on the balance error rate.

\end{itemize}

\section{Related Work}
\label{sec:related_work}

In this section, we focus on discussing works on single-image shadow detection rather than trying to be exhaustive.


Traditionally, single-image shadow detection is done by exploiting physical models of illumination and color~\cite{finlayson2006removal,finlayson2009entropy,tian2016new}.
This approach, however, tends to produce satisfactory results only for wide dynamic range images~\cite{lalonde2010detecting,nguyen2017shadow}.
Another approach learns shadow properties using hand-crafted features based on annotated shadow images.
It first describes image regions by feature descriptors and then classifies the regions into shadow and non-shadow regions.
Features like color~\cite{lalonde2010detecting,guo2011single,vicente2015leave}, texture~\cite{zhu2010learning,guo2011single,vicente2015leave}, edge~\cite{lalonde2010detecting,zhu2010learning,huang2011characterizes} and T-junction~\cite{lalonde2010detecting} are commonly used for shadow detection followed by classifiers like decision tree~\cite{lalonde2010detecting,zhu2010learning} and SVM~\cite{guo2011single,huang2011characterizes,vicente2015leave}. 
However, since hand-crafted features have limited capability in describing shadows, this approach often fails for complex cases.


\begin{figure*} [htbp]
	\centering
	\includegraphics[width=0.92\linewidth]{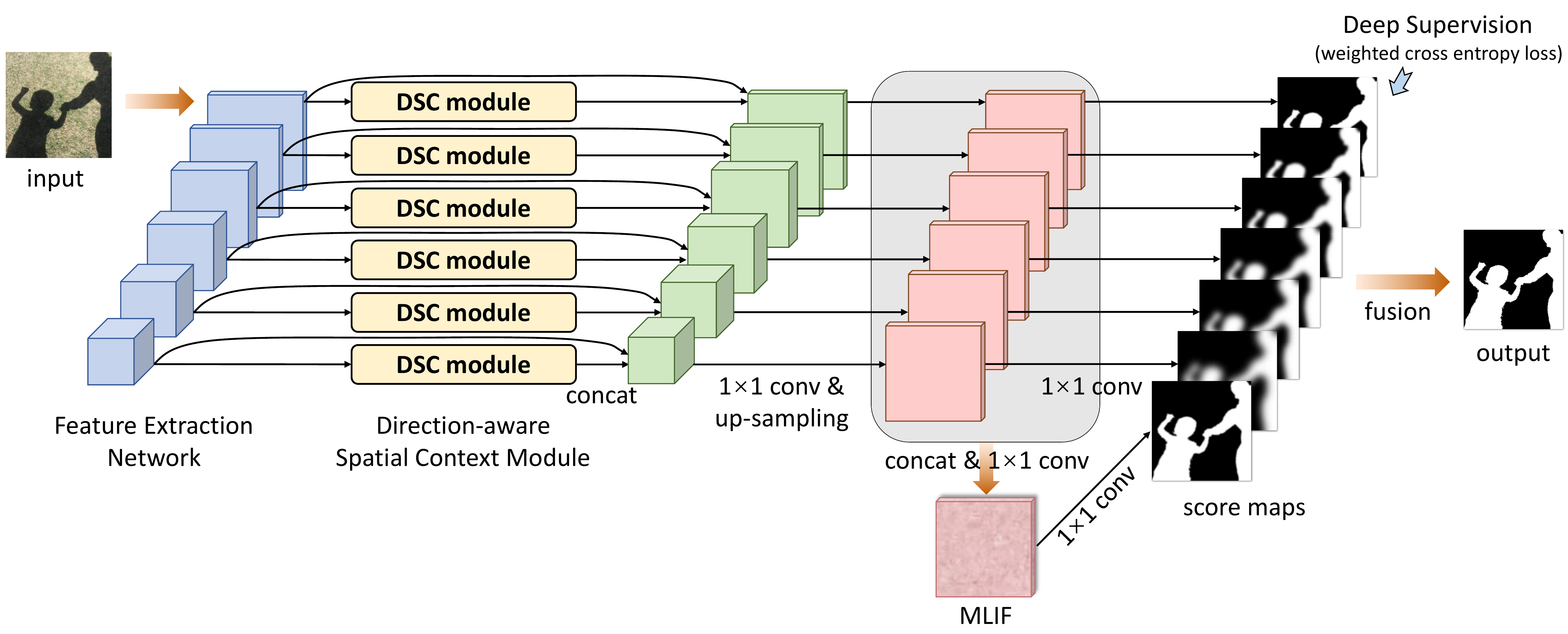}
	\caption{The schematic illustration of the overall shadow detection network:
    		 (i) we extract features in different scales over the CNN layers from the input image;
    		 (ii) we embed a DSC module (see Figure~\ref{fig:context}) to generate direction-aware spatial context (DSC) features for each layer;
    		 (iii) we concatenate the DSC features with convolutional features at each layer and upsample the concatenated feature maps to the size of the input image;
    		 (iv) we combine the upsampled feature maps into the multi-level integrated features (MLIF), and predict a score map based on the features for each layer by a deep supervision mechanism~\cite{xie2015holistically}; and
    		 (v) lastly, we fuse the resulting score maps to produce the final shadow detection result.}
	\label{fig:arc}
\end{figure*}


Convolutional neural network (CNN) is recently demonstrated to be a very powerful tool to learn features for detecting shadows, with results clearly outperforming previous approaches.
Khan et al.~\cite{khan2014automatic} used multiple CNNs to learn features in super pixels and along object boundaries, and fed the output features to a conditional random field to locate shadows.
Shen et al.~\cite{shen2015shadow} presented a deep structured shadow edge detector and employed structured labels to improve the local consistency of the predicted shadow map.
Vicente et al.~\cite{vicente2016large} trained stacked-CNN using a large data set with noisy annotations.
They minimized the sum of squared leave-one-out errors for image clusters to recover the annotations, and trained two CNNs to detect shadows.

Very recently, Hosseinzadeh et al.~\cite{hosseinzadeh2017fast} detected shadows using a patch-level CNN and a shadow prior map generated from hand-crafted features, while 
Nguyen et al.~\cite{nguyen2017shadow} developed scGAN with a sensitivity parameter to adjust weights in the loss functions.
Although the shadow detection accuracy keeps improving on the benchmark datasets~\cite{zhu2010learning,vicente2016large}, existing methods may still misrecognize black objects as shadows and miss unobvious shadows in the testing images.
The most recent work by Nguyen et al.~\cite{nguyen2017shadow} emphasized the importance of reasoning global semantics for shadow detection.
Compared to this work, we suggest to consider the directional variance when analyzing the spatial context.
Results show that our method can further outperform~\cite{nguyen2017shadow} in terms of both the accuracy and the BER value.


\section{Methodology}
\label{sec:method}

Figure~\ref{fig:arc} presents the workflow of the overall shadow detection network that employs the DSC module (see Figure~\ref{fig:context}) to learn direction-aware spatial context features.
Our network takes the whole image as input and outputs the shadow detection map in an end-to-end manner.
First, it begins by using a convolutional neural network (CNN) to extract a set of hierarchical feature maps, which encode fine details and semantic information in different scales over the CNN layers.
Second, for each layer, we employ a DSC module to harvest spatial contexts in a direction-aware manner and produce DSC features.
Third, we concatenate the DSC features with corresponding convolutional features, and upsample the concatenated feature map to the size of the input image.
Moreover, we further combine the upsampled feature maps into the multi-level integrated features (MLIF) with a convolution layer (via a $1\times1$ kernel), and apply the deep supervision mechanism~\cite{xie2015holistically} to impose the supervision signals to each layer as well as to the MLIF and predict a score map at each layer.
Lastly, we fuse all the predicted score maps into the final shadow map output.

In the following subsections, we first elaborate how the DSC module generates DSC features, and then introduce the training and testing strategies in the shadow detection network.


\subsection{Direction-aware Spatial Context}
\label{subsec:3.1}

\begin{figure*} [tp]
	\centering
	\includegraphics[width=0.98\linewidth]{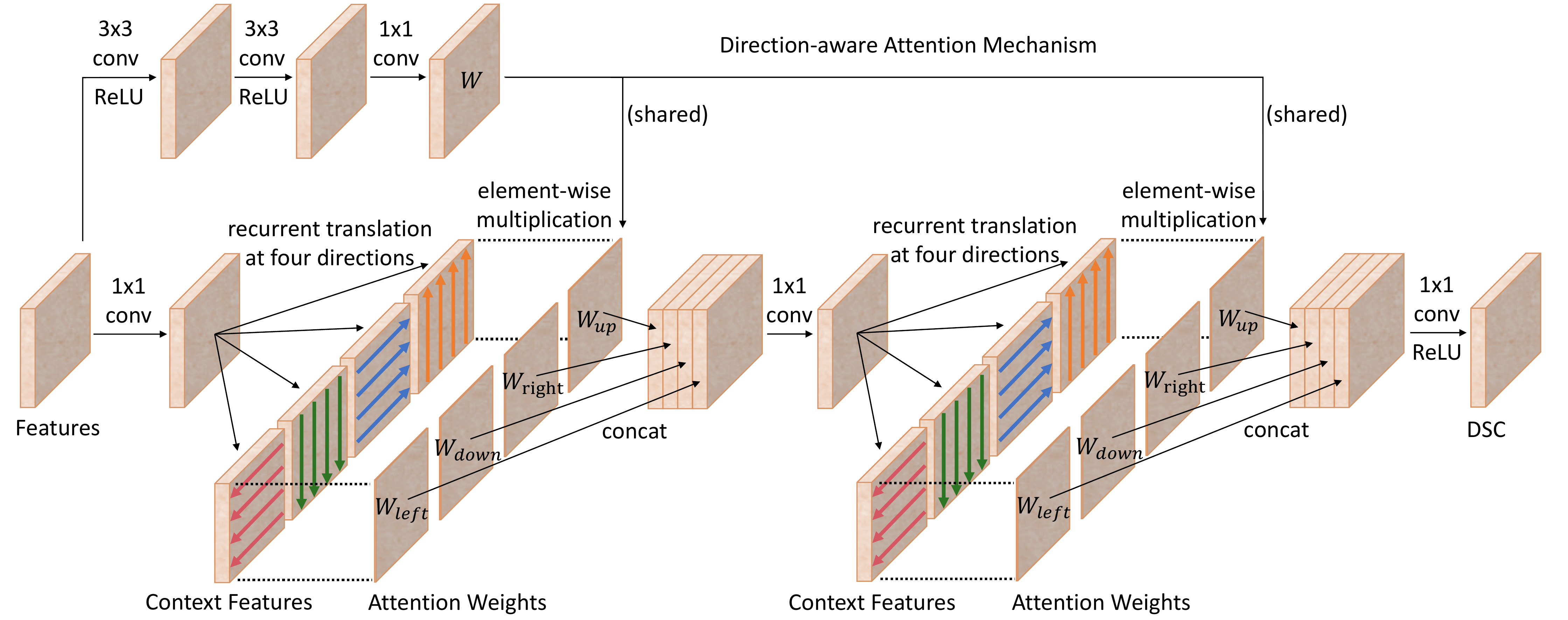}
	\caption{The schematic illustration of the {\em direction-aware spatial context module\/} ({\em DSC module\/}).
	We compute direction-aware spatial context by adopting a spatial RNN to aggregate spatial contexts in four principal directions with two rounds of recurrent translations, and formulate the attention mechanism to generate maps of attention weights to combine context features for different directions.
	We use the same set of weights in both rounds of recurrent translations.
	Best viewed in color.}
	\label{fig:context}	
\end{figure*}

Figure~\ref{fig:context} shows our DSC module architecture, which takes feature maps as input and outputs DSC features.
In this subsection, we first describe the concept of spatial context features and the spatial RNN model (Section~\ref{subsec:3.1.1}), and then elaborate how we formulate the direction-aware attention mechanism in a spatial RNN to learn attention weights and generate DSC features (Section~\ref{subsec:3.1.2}).


\subsubsection{Spatial Context Features}
\label{subsec:3.1.1}

\begin{figure} [tp]
	\centering
	\includegraphics[width=0.98\linewidth]{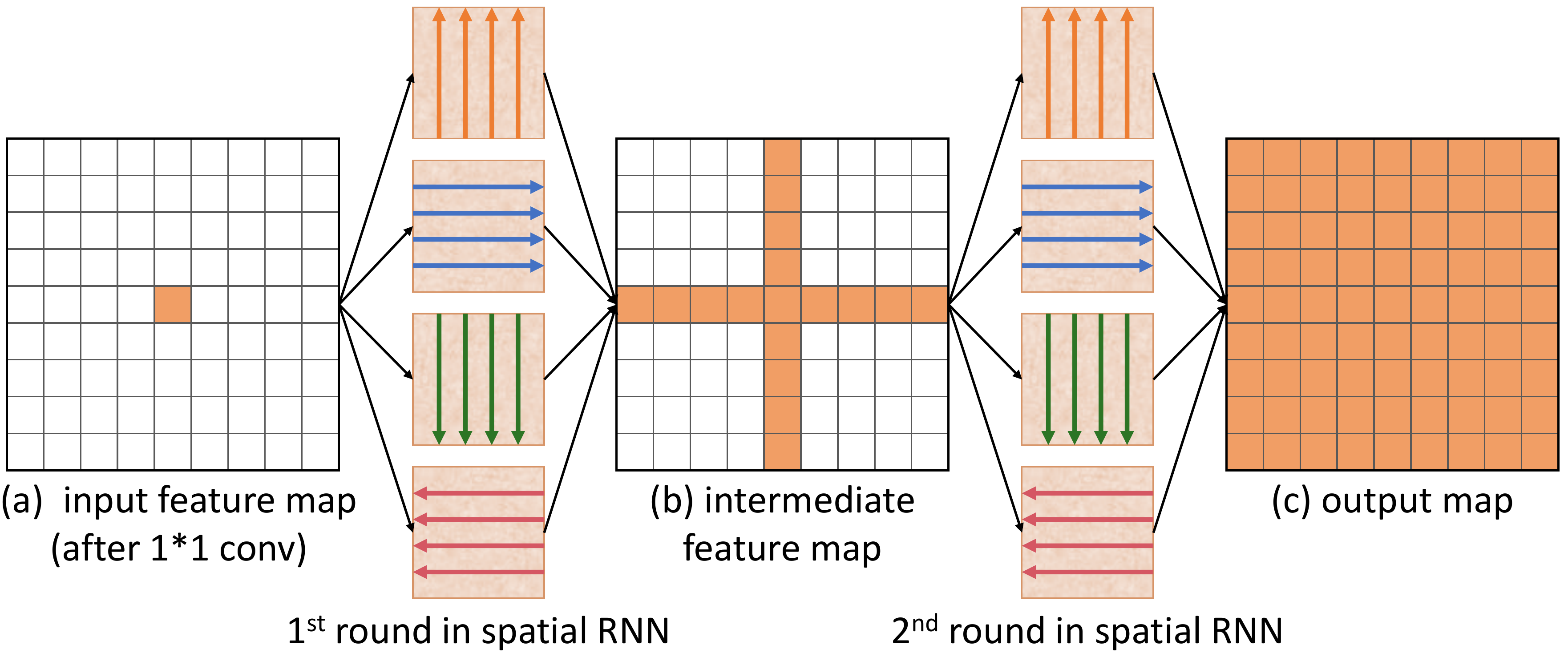}
	\caption{The schematic illustration of how spatial context information propagates in a two-round spatial RNN.}
	\label{fig:IRNN}
\end{figure}

Recurrent neural network (RNN)~\cite{lecun2015deep} is an effective model to process 1D sequential data via an array of input nodes (to receive data), an array of hidden nodes (to update internal states based on past and present data), and an array of output nodes (to output data).
There are three kinds of data translations in an RNN: from input nodes to hidden nodes, from hidden nodes to output nodes, and between adjacent hidden nodes.
By iteratively performing the data translations, the data received at input nodes can propagate across the hidden nodes, and eventually produce results at the output nodes.

For processing image data with 2D spatial context, RNN has been extended to build the spatial RNN model~\cite{bell2016inside}; see the schematic illustration in Figure~\ref{fig:IRNN}.
Taking a 2D feature map from a CNN as input, the spatial RNN model first uses a $1\times1$ convolution to perform an input-to-hidden data translation.
Then, it applies four independent data translations to aggregate local spatial context along each principal direction (left, right, up, and down), and fuses the results into an intermediate feature map; see Figure~\ref{fig:IRNN}(b).
Lastly, the whole process is repeated to further propagate the aggregated spatial context in each principal direction, and then to generate the overall spatial context; see Figure~\ref{fig:IRNN}(c).

Comparing with Figure~\ref{fig:IRNN}(c), each pixel in Figure~\ref{fig:IRNN}(a) knows only its local spatial context, while each pixel in Figure~\ref{fig:IRNN}(b) further knows the spatial context along the four principal directions after the first round of data translations.
Hence, by having two rounds of data translations, each pixel can obtain necessary global spatial context for learning features and solving the problem that the network is intended for.

To perform data translations in a spatial RNN, we follow the IRNN model~\cite{bell2016inside}, since it is fast, easy to train, and has a good performance for long-range data dependencies~\cite{bell2016inside}.
Denote $h_{i,j}$ as the feature at pixel $(i,j)$, we perform one round of data translations to the right (similarly for the other directions) by repeating the following operation $n$ times:
\begin{equation}  \label{recurrent}
h_{i,j} = \max( \ \alpha_\text{right} \ h_{i,j-1}+h_{i,j} \ , \ 0 \ )\ ,
\end{equation}
where $n$ is the width of the feature map and $\alpha_\text{right}$ is the weight parameter in the recurrent translation layer for the right direction.
Note that $\alpha_\text{right}$, as well as weights for the other directions, are initialized to be an identity matrix and are learned by the training process automatically.


\subsubsection{Direction-aware Spatial Context Features}
\label{subsec:3.1.2}

To learn spatial context in a direction-aware manner, we formulate the direction-aware attention mechanism in a spatial RNN to learn attention weights and generate direction-aware spatial context (DSC) features.

\paragraph{Direction-aware attention mechanism.}
The purpose of the direction-aware attention mechanism is to enable the spatial RNN to selectively leverage the spatial context aggregated along different directions by means of learning.
See the top-left blocks in the DSC module shown in Figure~\ref{fig:context}.
First, we employ two successive convolutional layers (with $3$$\times$$3$ kernels) followed by the ReLU~\cite{krizhevsky2012imagenet} non-linear operation, and then the third convolutional layer (with $1$$\times$$1$ kernels) to generate $\mathbf{W}$.
Then, we split $\mathbf{W}$ into four maps of attention weights denoted as $\mathbf{W}_\text{left}$, $\mathbf{W}_\text{down}$, $\mathbf{W}_\text{right}$, and $\mathbf{W}_\text{up}$.
Mathematically, if we denote the above operators as $f_{att}$ and the input feature maps as $\mathbf{X}$, we have
\begin{equation}
\label{attention}
\mathbf{W} \ = \ f_{att}( \ \mathbf{X} \ ; \ \theta \ )\ ,
\end{equation}
where $\theta$ denotes the parameters to be learned by $f_{att}$, and $f_{att}$ is also known as the attention estimator network.

See again the DSC module shown in Figure~\ref{fig:context}.
The four maps of weights are multiplied with the spatial context features (from the recurrent data translations) along different directions in an element-wise manner.
Therefore, after we train the network with the shadow dataset, the network can learn $\theta$ for producing suitable attention weights to selectively leverage the spatial context in the spatial RNN.


\paragraph{Completing the DSC module.}
Next, we further provide additional details about the DSC module.
As shown in Figure~\ref{fig:context}, after we multiply the spatial context features with the attention weights, we concatenate the results and use a $1\times1$ convolution to simulate a hidden-to-hidden data translation and reduce the feature dimensions to a quarter of the dimension size.
Then, we perform the second round of recurrent translations and use the same set of attention weights to select spatial context.
We empricially find that the network delivers higher performance, if we share the attention weights rather than using two separate sets of weights.
Note that these attention weights are learnt based on the deep features extracted from the input images, and they may vary from images to images.
Lastly, we utilize a $1\times1$ convolution followed by the ReLU~\cite{krizhevsky2012imagenet} non-linear operation on the concatenated feature maps to simulate the hidden-to-output translation and produce the output DSC features.


\subsection{Training and Testing Strategies}
\label{subsec:3.2}

Our network is built upon the VGG network~\cite{simonyan2014very}, where we apply a DSC module to each layer, except for the first layer, which involves a large memory footprint.

\paragraph{Loss Function.}
In natural images, shadows usually occupy smaller areas than non-shadow regions.
Hence, if the loss function simply aims for overall accuracy, it will incline to match the non-shadow regions, which have far more pixels.
Therefore, we use a weighted cross-entropy loss to optimize the whole network in the training process.

In detail, assume that the ground truth value of a pixel is $y$ (where $y$$=$$1$, if it is in shadow, and $y$$=$$0$, otherwise) and the prediction label of the pixel is $p$ (where $p\in[0,1]$).
The weighted cross entropy loss $L$ equals $L_1 + L_2$:
\begin{equation}  \label{loss1}
L_1 = -(\frac{N_n}{N_p+N_n}) y \log(p) - (\frac{N_p}{N_p+N_n})(1 - y) \log(1 - p)\ ,
\end{equation}
and
\begin{equation}  \label{loss2}
L_2 = -(1-\frac{TP}{N_p}) y \log(p) - (1-\frac{TN}{N_n})(1 - y) \log(1 - p)\ ,
\end{equation}
where $TP$ and $TN$ are the number of true positives and true negatives, and $N_p$ and $N_n$ are the number of shadow and non-shadow pixels, respectively, so $N_p$$+$$N_n$ is the total number of pixels.
In practice, $L_1$ helps balance the detection of shadows and non-shadows; if the area of shadows is less than that of the non-shadow region, we will penalize misclassified shadow pixels more than misclassified non-shadow pixels.
On the other hand, $L_2$ helps the network focus on learning the class (shadow or non-shadow) that is difficult to be classified~\cite{shrivastava2016training}.
This can be achieved, since the weight in loss function for shadow (or non-shadow) class is large when the number of correctly-classified shadow (or non-shadow) pixels is small, and vice versa.

We use the above loss function for each layer in the shadow detection network presented in Figure~\ref{fig:arc}.
Hence, the overall loss function $L_\text{overall}$ is a summation of the individual loss on all the predicted score maps:
\begin{equation}
\label{loss_t}
L_\text{overall} \ = \ \sum_{i} w_i L_i \ + \ w_m L_m \ + \ w_f L_f \ ,
\end{equation}
where $w_i$ and $L_i$ denote the weight and loss of the $i$-th layer (level) in the overall network, respectively;
$w_m$ and $L_m$ are the weight and loss of the MLIF layer; and
$w_f$ and $L_f$ are the weight and loss of the fusion layer, which is the last layer in the overall network to produce the final shadow detection result; see Figure~\ref{fig:arc}.
Note that all the weights $w_i$, $w_m$ and $w_f$ are empirically set to be $1$.


\paragraph{Training parameters.}
To accelerate the training process while reducing over-fitting, we initialize parameters in the feature extraction layers (see the frontal part of the network in Figure~\ref{fig:arc}) by the well-trained VGG network~\cite{simonyan2014very} and the parameters in other layers by random noise.
Stochastic gradient descent is used to optimize the whole network with a momentum value of $0.9$ and a weight decay of $5$$\times$$10^{-4}$.
We set the learning rate as $10^{-8}$ and terminate the learning process after 12k iterations.
Moreover, we horizontally flip images for data argumentation.
Note that we build the model on Caffe~\cite{jia2014caffe} with a mini-batch size of 1.


\paragraph{Inference.}

In the testing process, our network produces one score map for each layer, including the MLIF layer and the fusion layer, with a supervision signal added to each layer.
After that, we compute the mean of the score maps over the MLIF layer and the fusion layer to produce the final prediction map.
Lastly, we apply the fully connected conditional field~\cite{krahenbuhl2011efficient} to improve the detection result by considering the spatial coherence between neighborhood pixels.

\section{Experimental Results}
\label{sec:experiments}

\begin{figure*}[t]
	\centering

	\vspace*{0.5mm}
	\begin{subfigure}{0.107\textwidth} 
		\includegraphics[width=\textwidth]{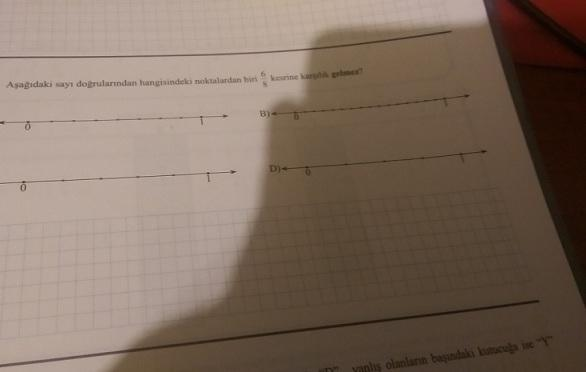}
	\end{subfigure}
	\begin{subfigure}{0.107\textwidth}
		\includegraphics[width=\textwidth]{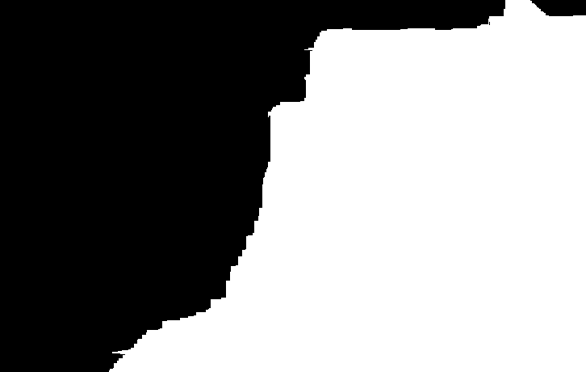}
	\end{subfigure}
	\begin{subfigure}{0.107\textwidth}
		\includegraphics[width=\textwidth]{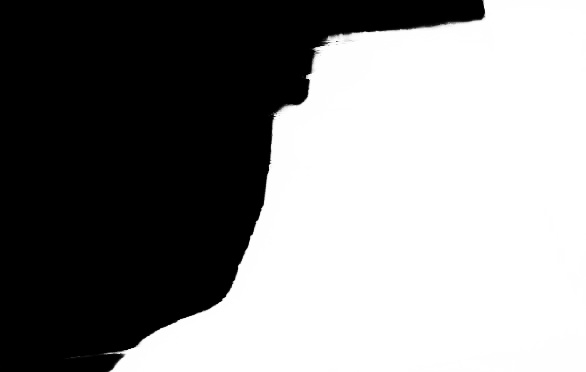}
	\end{subfigure}
	\begin{subfigure}{0.107\textwidth}
		\includegraphics[width=\textwidth]{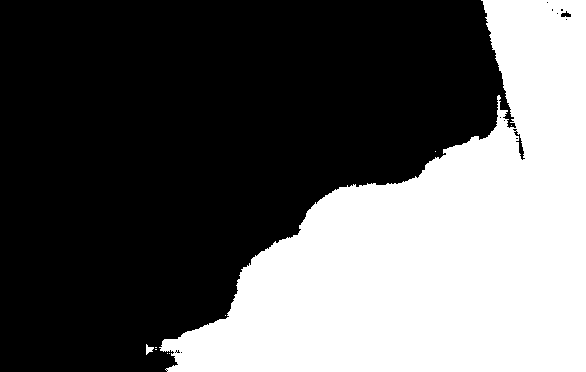}
	\end{subfigure}
	\begin{subfigure}{0.107\textwidth}
		\includegraphics[width=\textwidth]{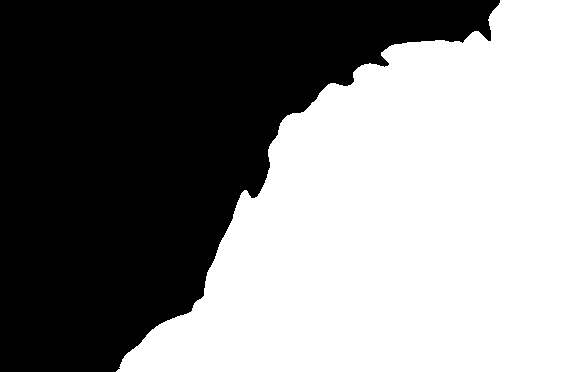}
	\end{subfigure}
	\begin{subfigure}{0.107\textwidth}
		\includegraphics[width=\textwidth]{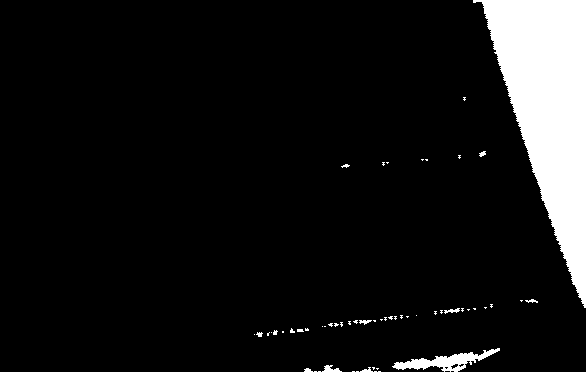}
	\end{subfigure}
	\begin{subfigure}{0.107\textwidth}
		\includegraphics[width=\textwidth]{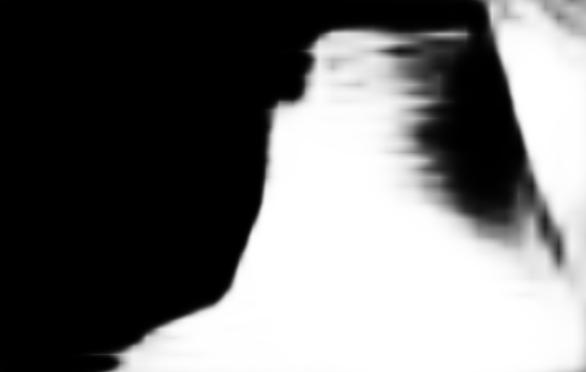}
	\end{subfigure}
	\begin{subfigure}{0.107\textwidth}
		\includegraphics[width=\textwidth]{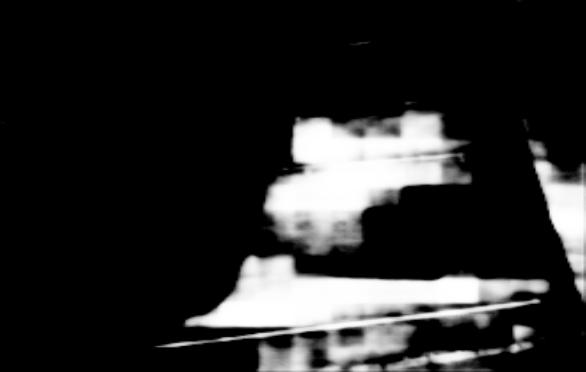}
	\end{subfigure}
	\begin{subfigure}{0.107\textwidth}
		\includegraphics[width=\textwidth]{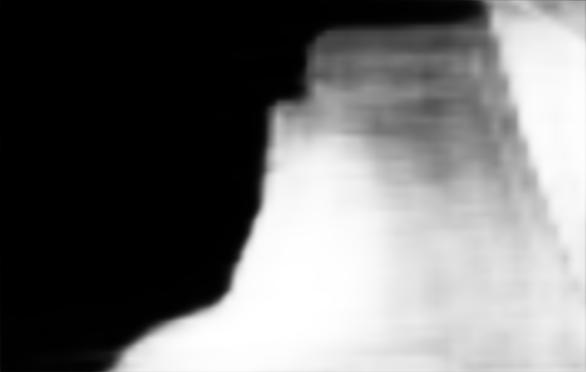}
	\end{subfigure}
	
	\ \\
	
	\vspace*{0.5mm}
	\begin{subfigure}{0.107\textwidth}
		\includegraphics[width=\textwidth]{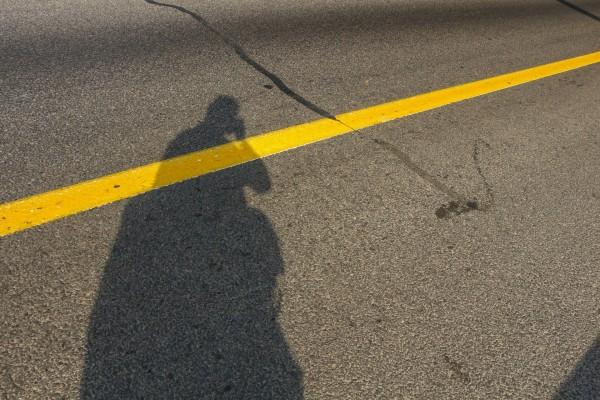}
	\end{subfigure}
	\begin{subfigure}{0.107\textwidth}
		\includegraphics[width=\textwidth]{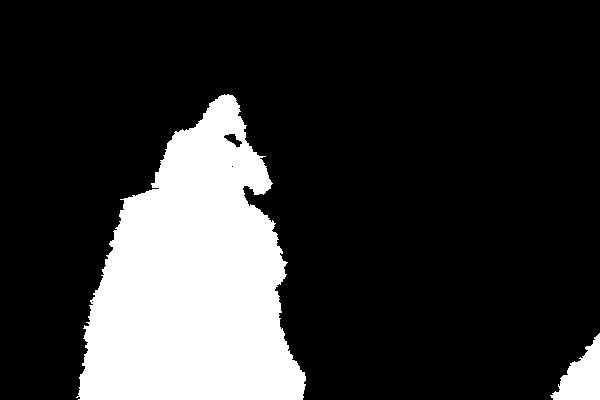}
	\end{subfigure}
	\begin{subfigure}{0.107\textwidth}
		\includegraphics[width=\textwidth]{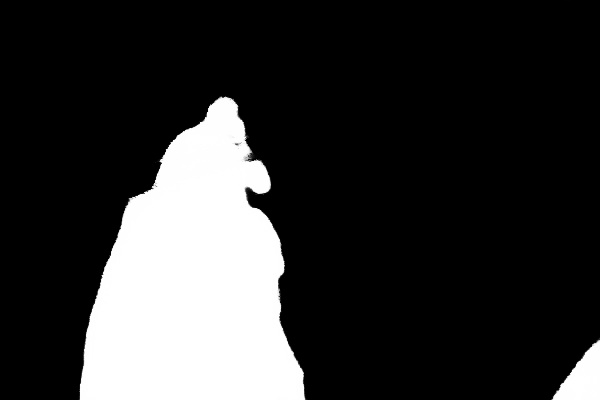}
	\end{subfigure}
	\begin{subfigure}{0.107\textwidth}
		\includegraphics[width=\textwidth]{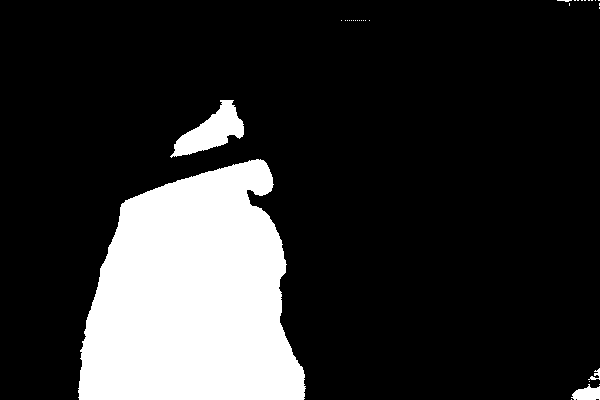}
	\end{subfigure}
	\begin{subfigure}{0.107\textwidth}
		\includegraphics[width=\textwidth]{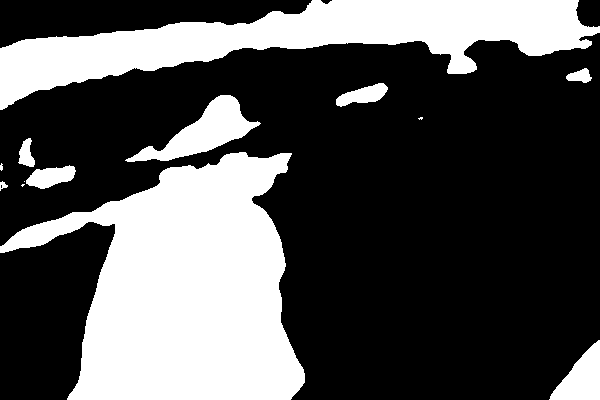}
	\end{subfigure}
	\begin{subfigure}{0.107\textwidth}
		\includegraphics[width=\textwidth]{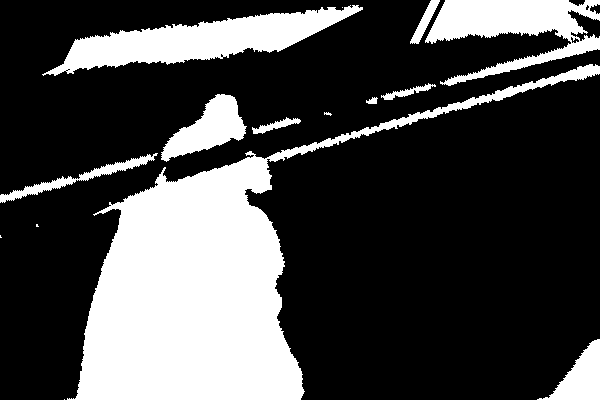}
	\end{subfigure}
	\begin{subfigure}{0.107\textwidth}
		\includegraphics[width=\textwidth]{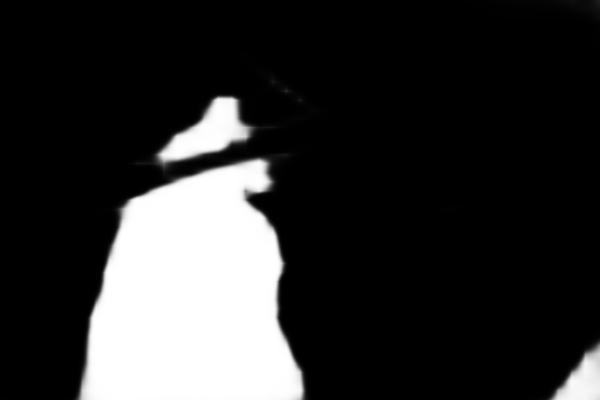}
	\end{subfigure}
	\begin{subfigure}{0.107\textwidth}
		\includegraphics[width=\textwidth]{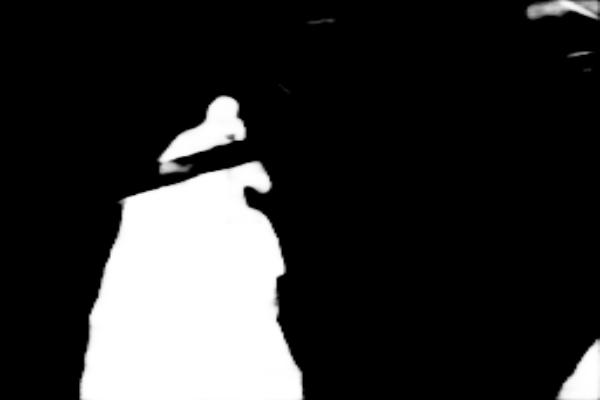}
	\end{subfigure}
	\begin{subfigure}{0.107\textwidth}
		\includegraphics[width=\textwidth]{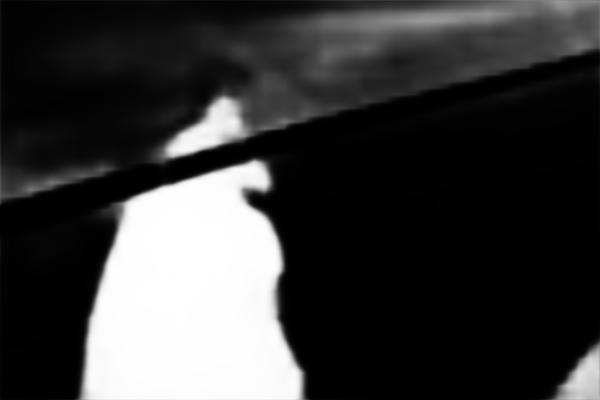}
	\end{subfigure}
	
	\ \\
	
	\vspace*{0.5mm}
	\begin{subfigure}{0.107\textwidth}
		\includegraphics[width=\textwidth]{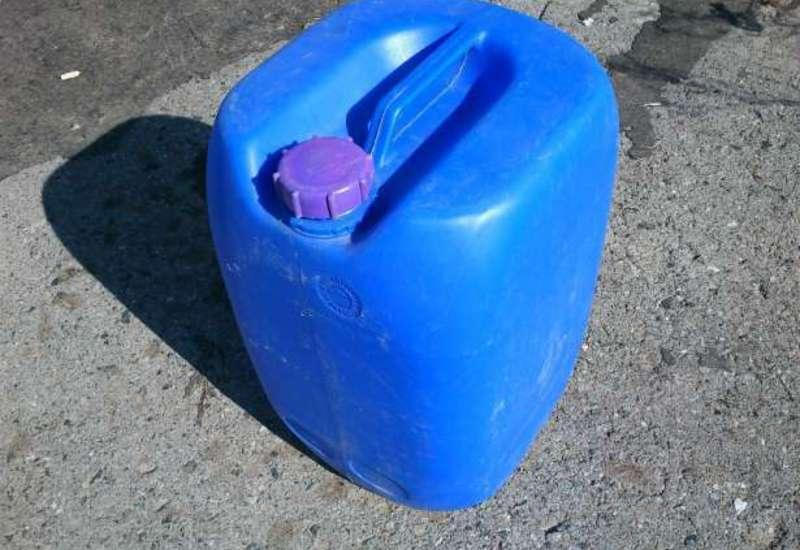}
	\end{subfigure}
	\begin{subfigure}{0.107\textwidth}
		\includegraphics[width=\textwidth]{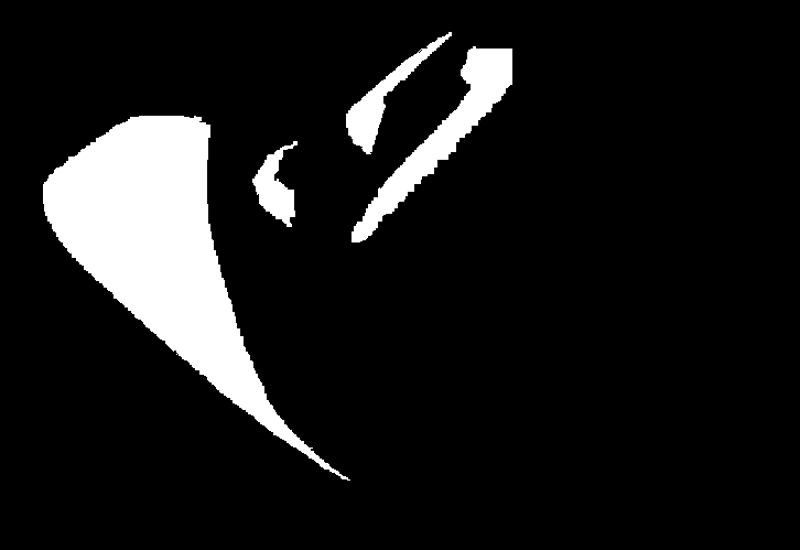}
	\end{subfigure}
	\begin{subfigure}{0.107\textwidth}
		\includegraphics[width=\textwidth]{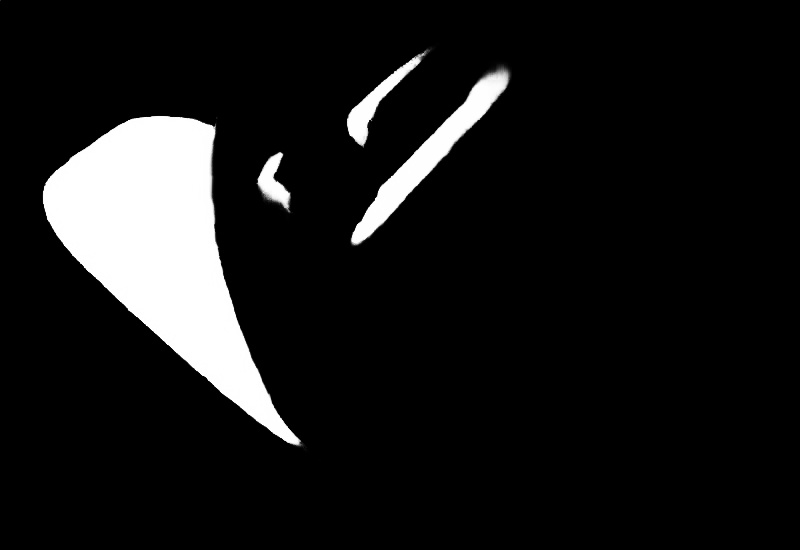}
	\end{subfigure}
	\begin{subfigure}{0.107\textwidth}
		\includegraphics[width=\textwidth]{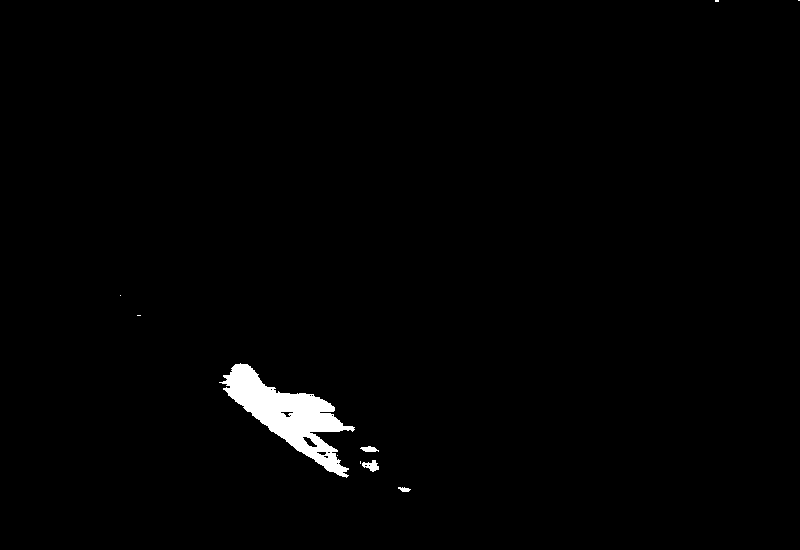}
	\end{subfigure}
	\begin{subfigure}{0.107\textwidth}
		\includegraphics[width=\textwidth]{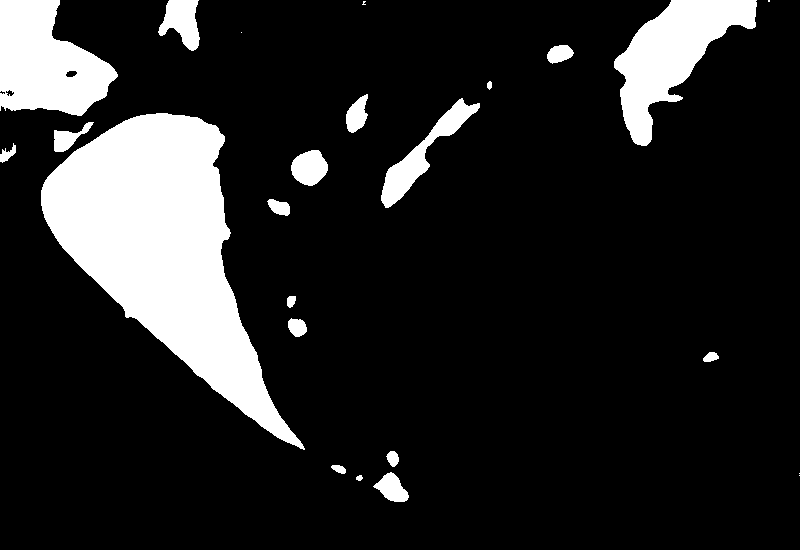}
	\end{subfigure}
	\begin{subfigure}{0.107\textwidth}
		\includegraphics[width=\textwidth]{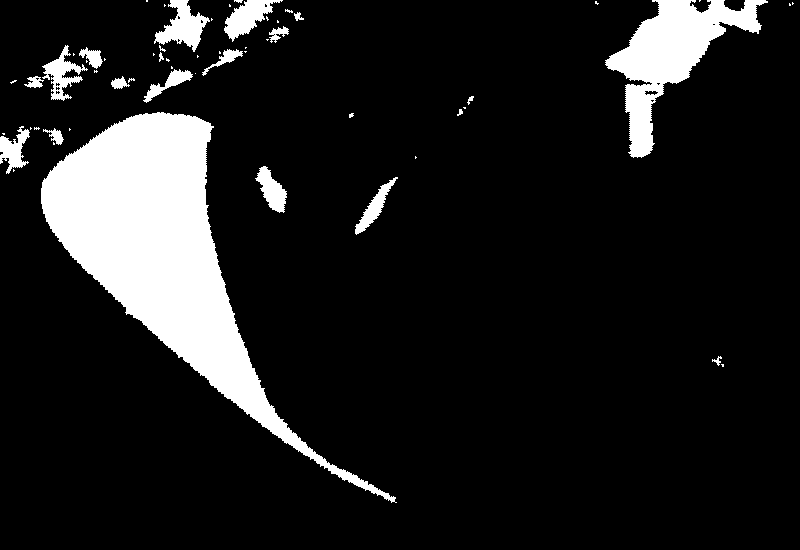}
	\end{subfigure}
	\begin{subfigure}{0.107\textwidth}
		\includegraphics[width=\textwidth]{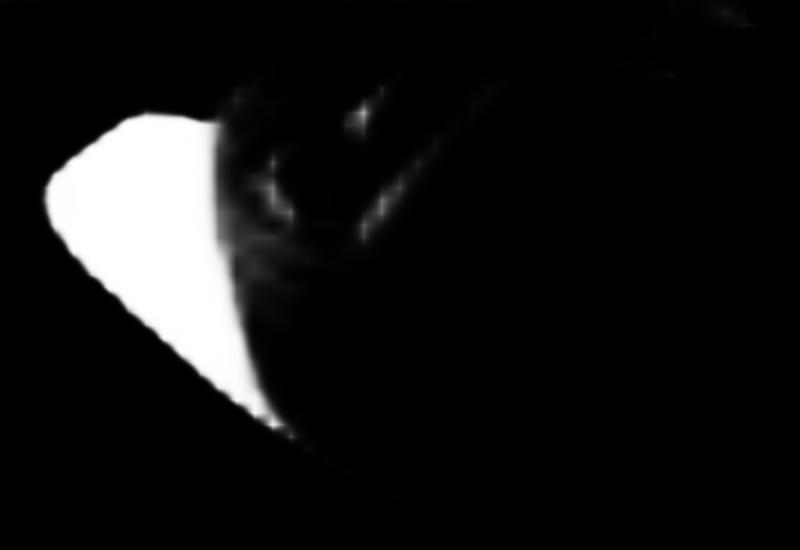}
	\end{subfigure}
	\begin{subfigure}{0.107\textwidth}
		\includegraphics[width=\textwidth]{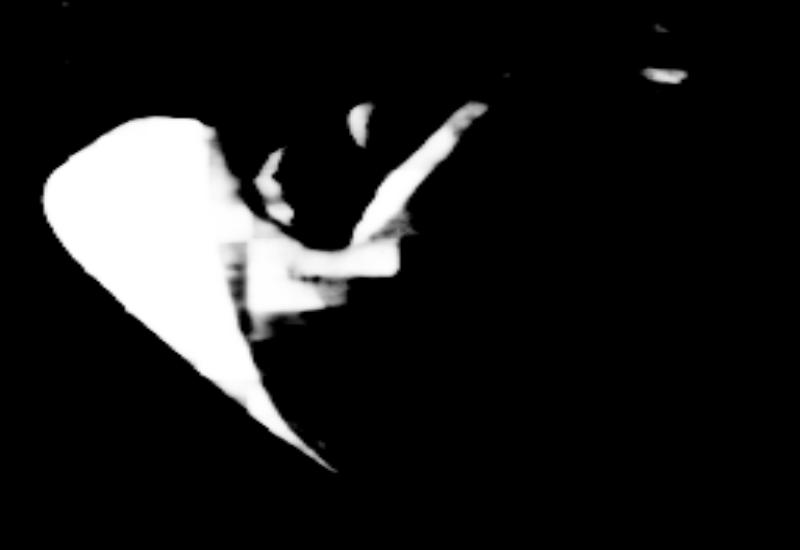}
	\end{subfigure}
	\begin{subfigure}{0.107\textwidth}
		\includegraphics[width=\textwidth]{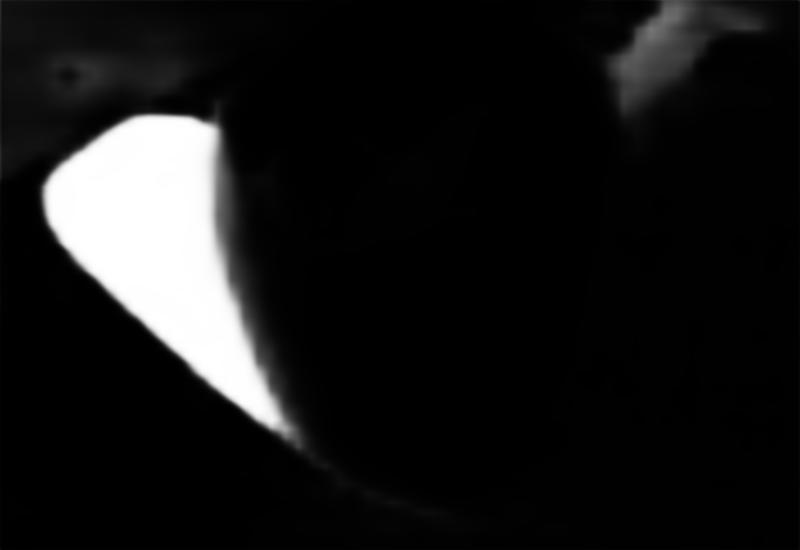}
	\end{subfigure}
	
	\ \\
	
	\vspace*{0.5mm}
	\begin{subfigure}{0.107\textwidth}
		\includegraphics[width=\textwidth]{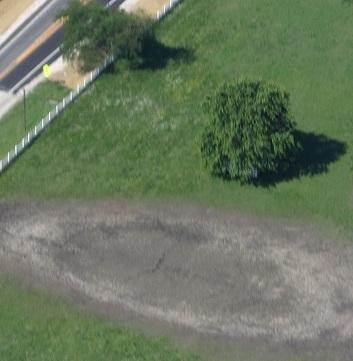}
	\end{subfigure}
	\begin{subfigure}{0.107\textwidth}
		\includegraphics[width=\textwidth]{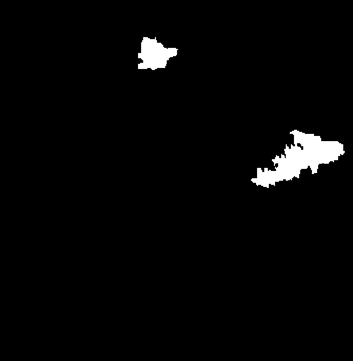}
	\end{subfigure}
	\begin{subfigure}{0.107\textwidth}
		\includegraphics[width=\textwidth]{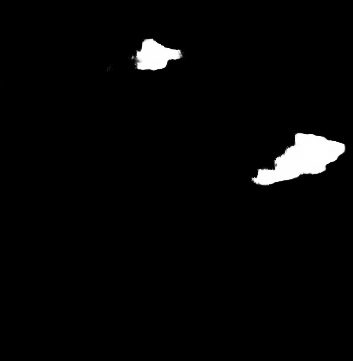}
	\end{subfigure}
	\begin{subfigure}{0.107\textwidth}
		\includegraphics[width=\textwidth]{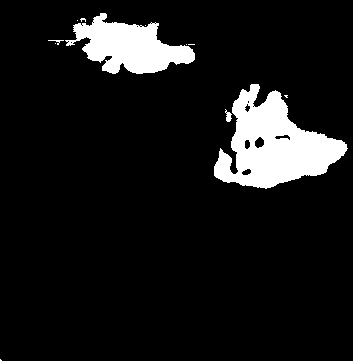}
	\end{subfigure}
	\begin{subfigure}{0.107\textwidth}
		\includegraphics[width=\textwidth]{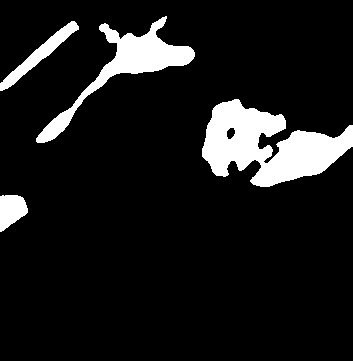}
	\end{subfigure}
	\begin{subfigure}{0.107\textwidth}
		\includegraphics[width=\textwidth]{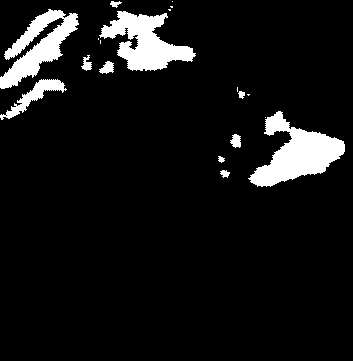}
	\end{subfigure}
	\begin{subfigure}{0.107\textwidth}
		\includegraphics[width=\textwidth]{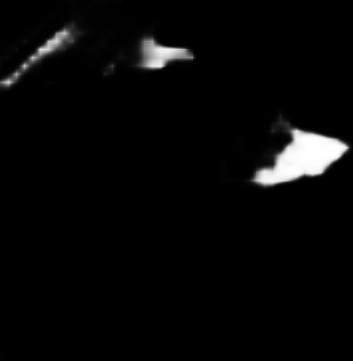}
	\end{subfigure}
	\begin{subfigure}{0.107\textwidth}
		\includegraphics[width=\textwidth]{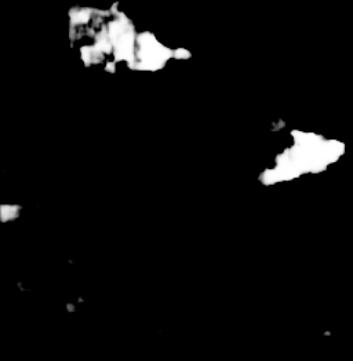}
	\end{subfigure}
	\begin{subfigure}{0.107\textwidth}
		\includegraphics[width=\textwidth]{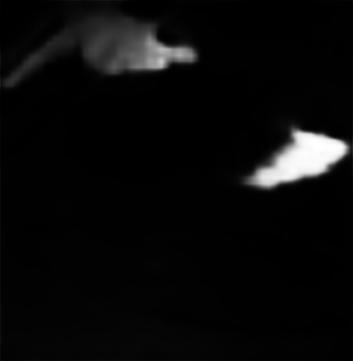}
	\end{subfigure}
	
	\ \\
	
	\vspace*{0.5mm}
	\begin{subfigure}{0.107\textwidth} 
		\includegraphics[width=\textwidth]{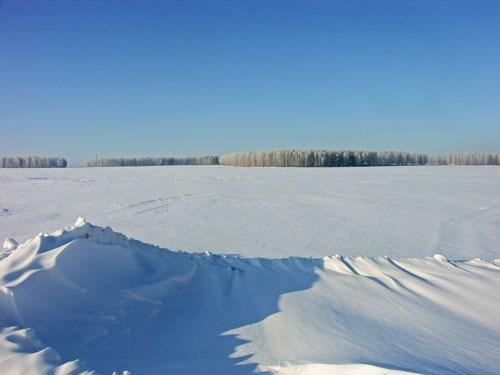}
	\end{subfigure}
	\begin{subfigure}{0.107\textwidth}
		\includegraphics[width=\textwidth]{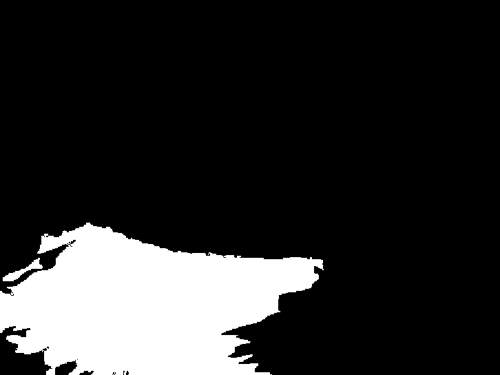}
	\end{subfigure}
	\begin{subfigure}{0.107\textwidth}
		\includegraphics[width=\textwidth]{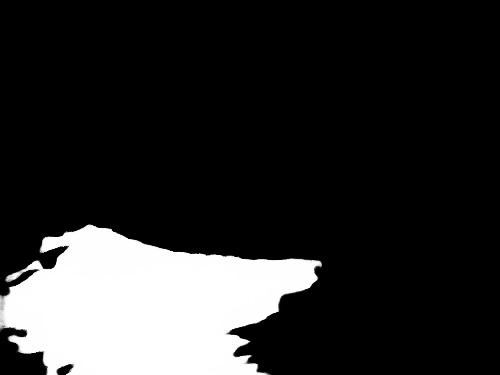}
	\end{subfigure}
	\begin{subfigure}{0.107\textwidth}
		\includegraphics[width=\textwidth]{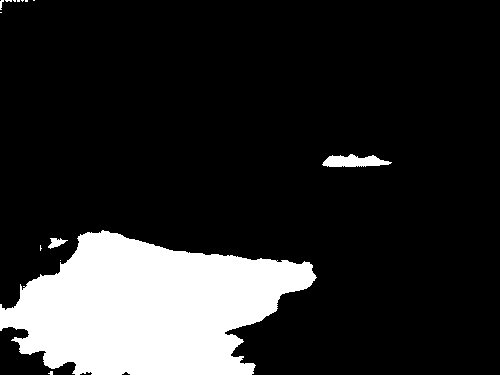}
	\end{subfigure}
	\begin{subfigure}{0.107\textwidth}
		\includegraphics[width=\textwidth]{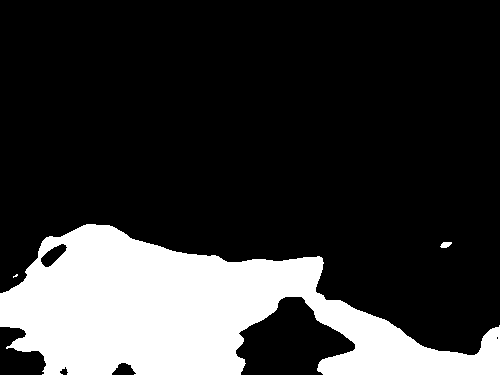}
	\end{subfigure}
	\begin{subfigure}{0.107\textwidth}
		\includegraphics[width=\textwidth]{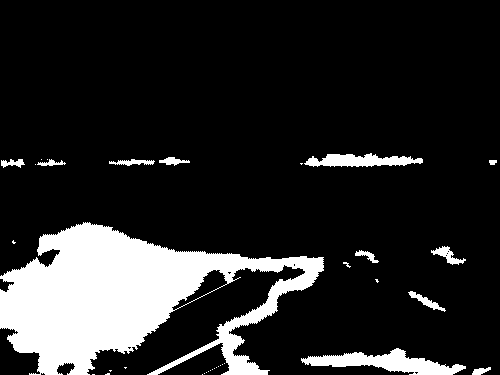}
	\end{subfigure}
	\begin{subfigure}{0.107\textwidth}
		\includegraphics[width=\textwidth]{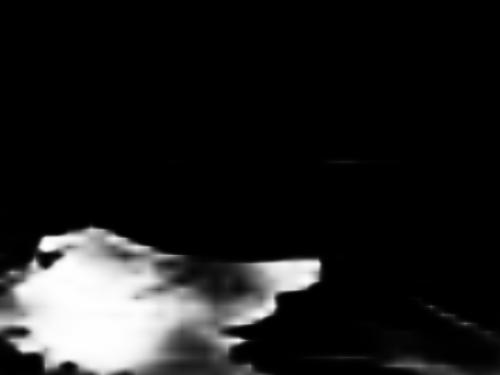}
	\end{subfigure}
	\begin{subfigure}{0.107\textwidth}
		\includegraphics[width=\textwidth]{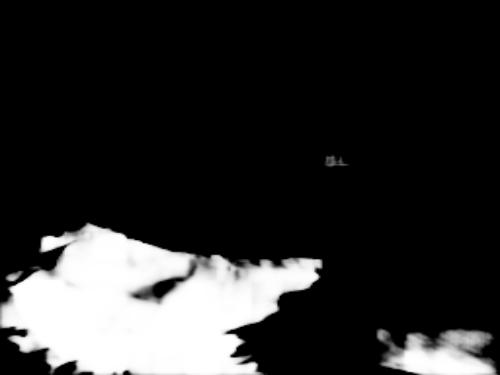}
	\end{subfigure}
	\begin{subfigure}{0.107\textwidth}
		\includegraphics[width=\textwidth]{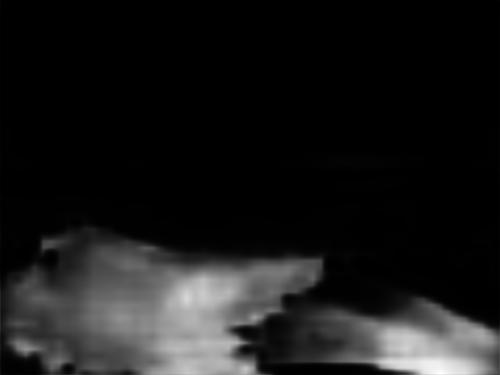}
	\end{subfigure}
	
	\ \\
	
	\vspace*{0.5mm}
	\begin{subfigure}{0.107\textwidth}
		\includegraphics[width=\textwidth]{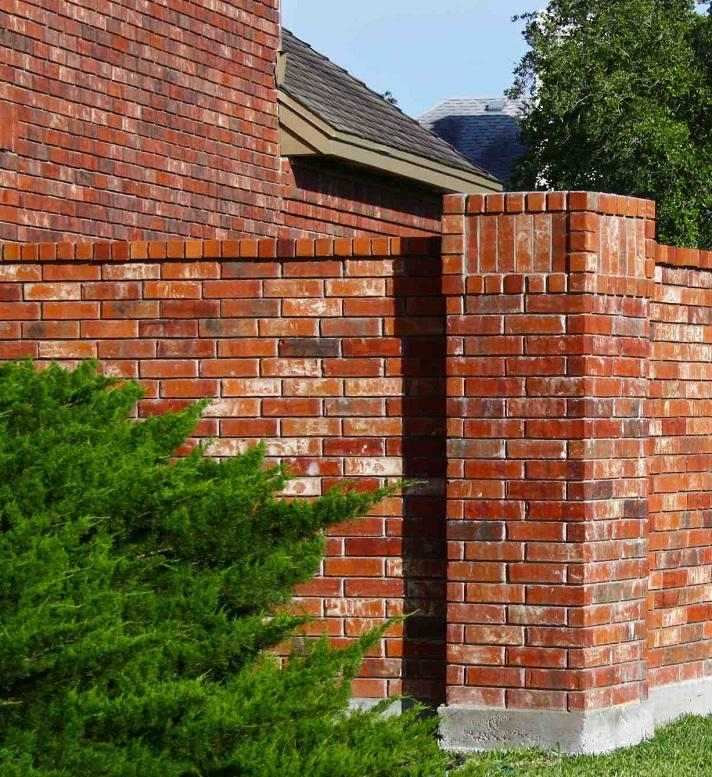}
		\vspace{-5.5mm} \caption*{{\footnotesize input image}}
	\end{subfigure}
	\begin{subfigure}{0.107\textwidth}
		\includegraphics[width=\textwidth]{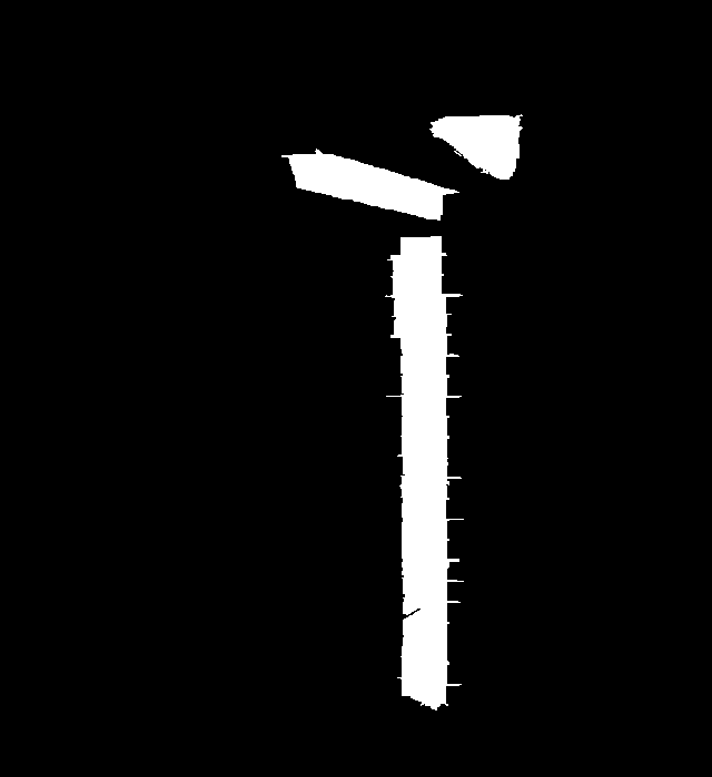}
		\vspace{-5.5mm} \caption*{{\footnotesize ground truth}} 
	\end{subfigure}
	\begin{subfigure}{0.107\textwidth}
		\includegraphics[width=\textwidth]{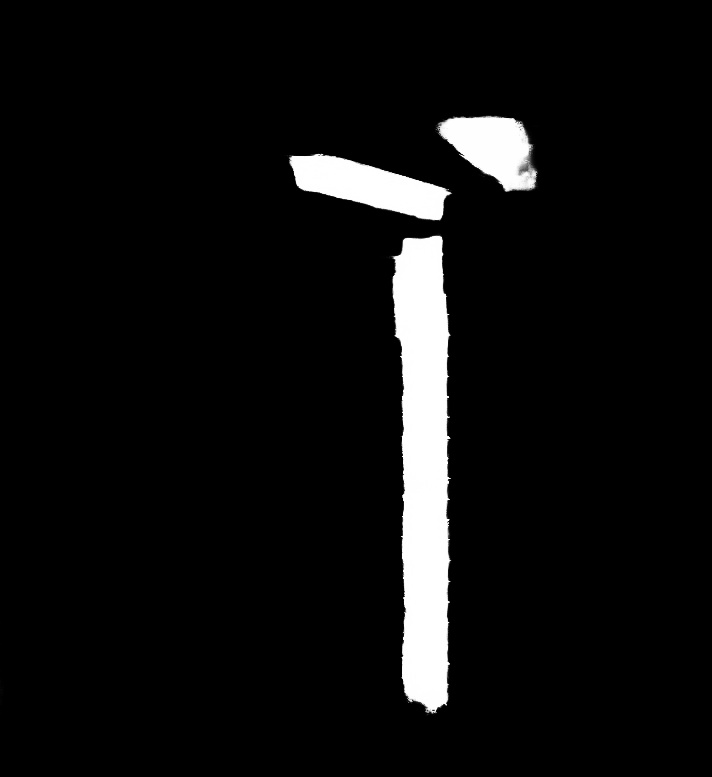}
		\vspace{-5.5mm} \caption*{{\footnotesize DSC (ours)}} 
	\end{subfigure}
	\begin{subfigure}{0.107\textwidth}
		\includegraphics[width=\textwidth]{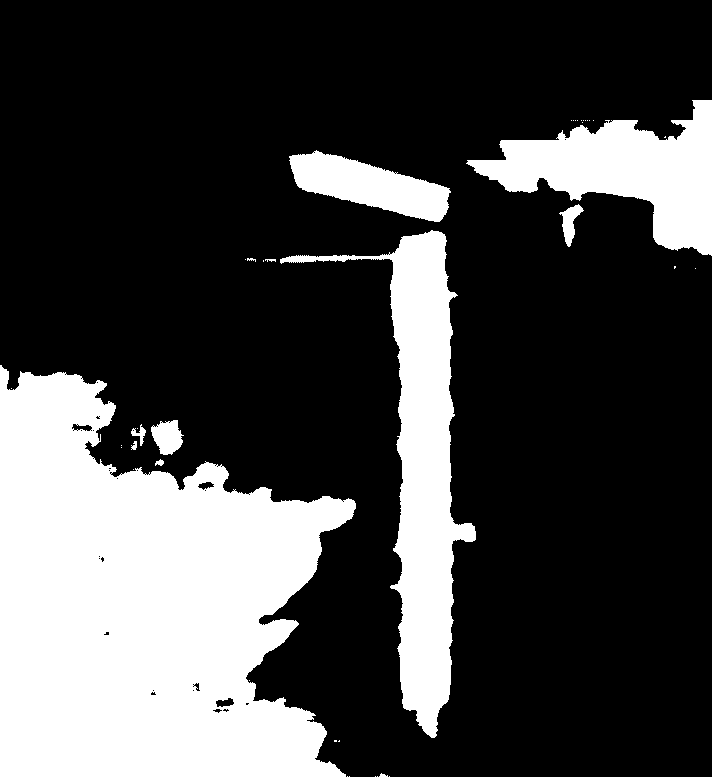}
		\vspace{-5.5mm} \caption*{{\footnotesize scGAN~\cite{nguyen2017shadow}}}
	\end{subfigure}
	\begin{subfigure}{0.107\textwidth}
		\includegraphics[width=\textwidth]{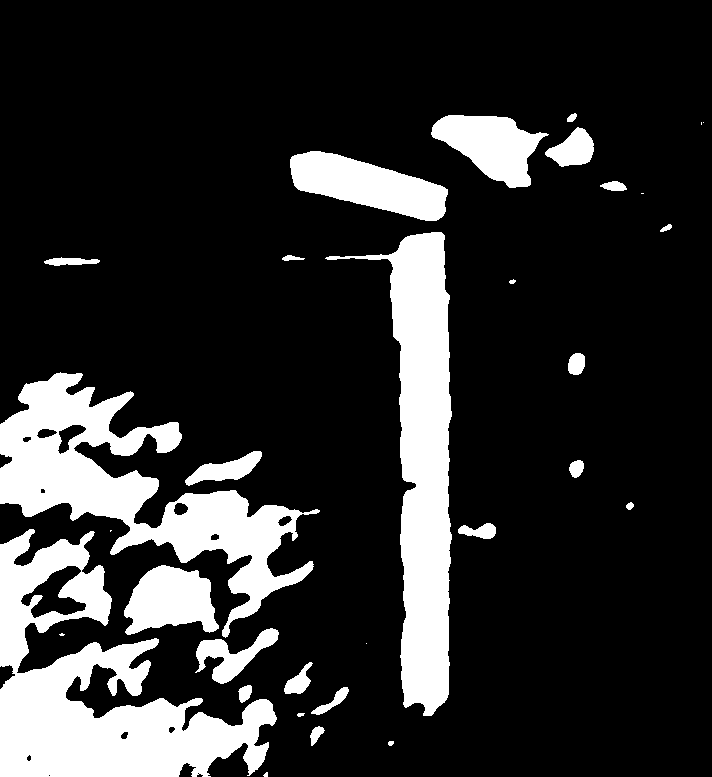}
		\vspace{-5.5mm} \caption*{{\footnotesize stkd'-CNN~\cite{vicente2016large}}}
	\end{subfigure}
	\begin{subfigure}{0.107\textwidth}
		\includegraphics[width=\textwidth]{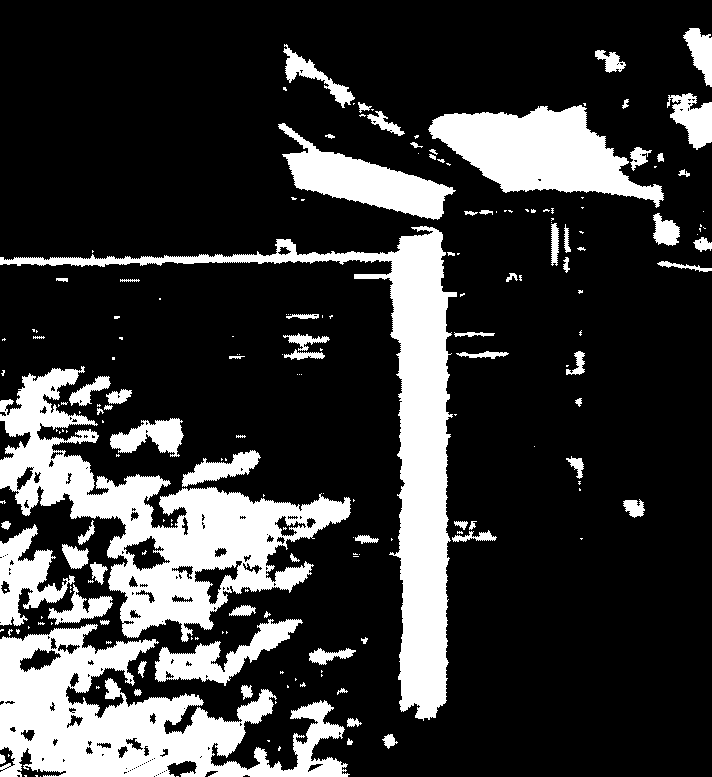}
		\vspace{-5.5mm} \caption*{{\footnotesize patd'-CNN~\cite{hosseinzadeh2017fast}}}
	\end{subfigure}
	\begin{subfigure}{0.107\textwidth}
		\includegraphics[width=\textwidth]{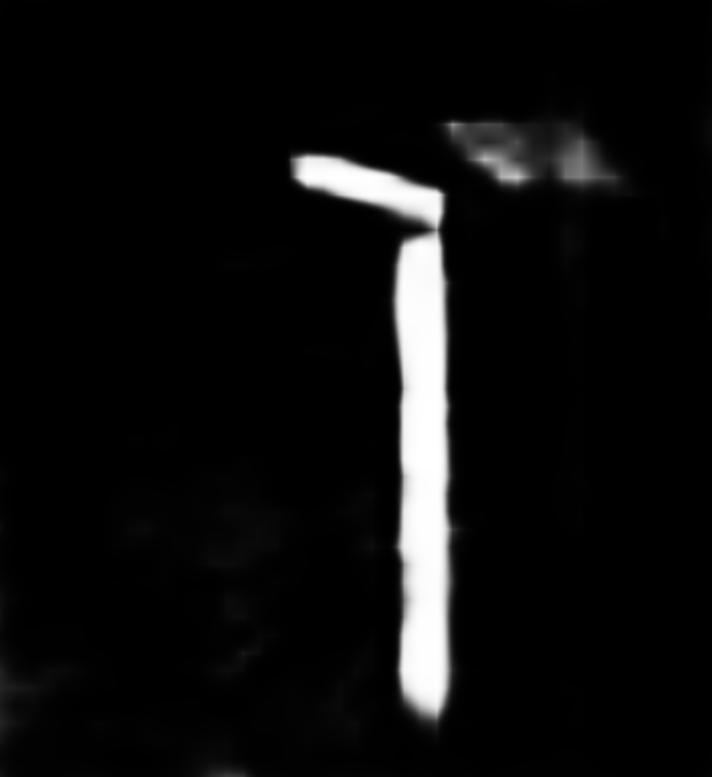}
		\vspace{-5.5mm} \caption*{{\footnotesize SRM~\cite{wang2017stagewise}}}
	\end{subfigure}
	\begin{subfigure}{0.107\textwidth}
		\includegraphics[width=\textwidth]{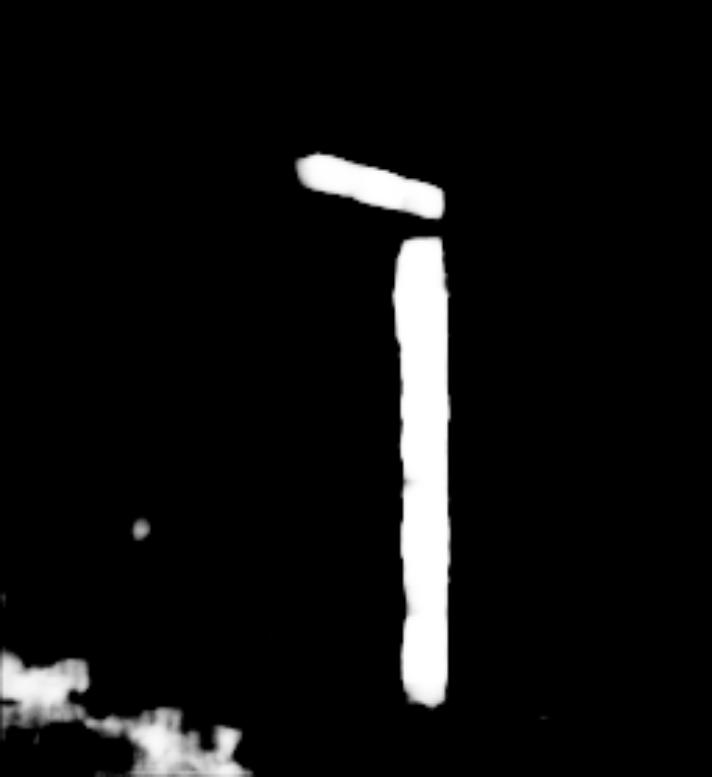}
		\vspace{-5.5mm} \caption*{{\footnotesize Amulet~\cite{zhang2017amulet}}}
	\end{subfigure}
	\begin{subfigure}{0.107\textwidth}
		\includegraphics[width=\textwidth]{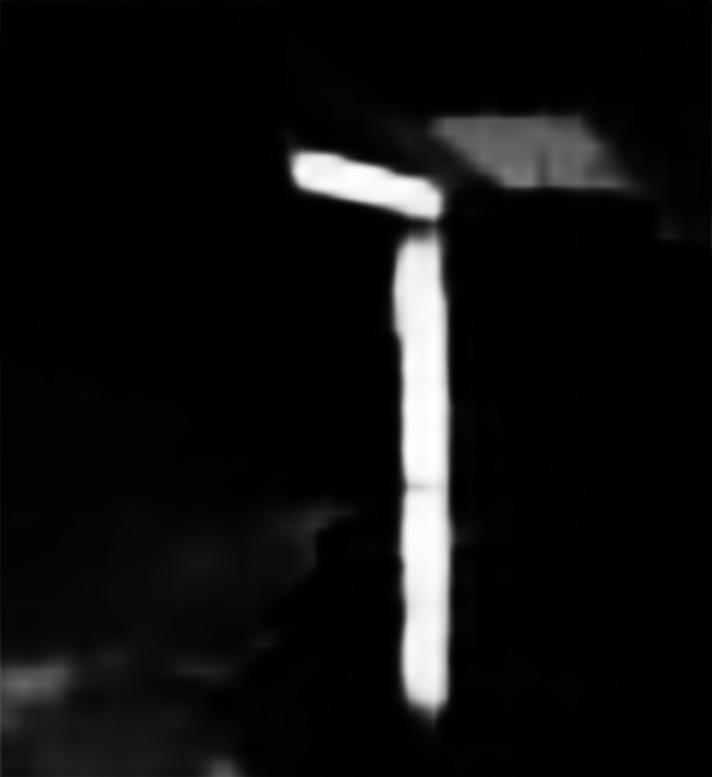}
		\vspace{-5.5mm} \caption*{\hspace*{-0.7mm}{\footnotesize PSPNet~\cite{Zhao_2017_CVPR}}}
	\end{subfigure}
	
	\vspace{-1.0mm}
	\caption{Visual comparison of shadow maps produced by our method and other methods (4th-9th columns) against ground truths shown in 2nd column.
		Note that stkd'-CNN and patd'-CNN stand for stacked-CNN and patched-CNN, respectively.}
	
	\label{fig:comparison_real_photos}
	\vspace{-2.5mm}
\end{figure*}


\subsection{Datasets and Evaluation Metrics}


\paragraph{Benchmark datasets.}
Two benchmark datasets are employed in this work.
The first one is the SBU Shadow Dataset~\cite{vicente2016large}, which is the largest publicly available annotated shadow dataset with 4089 training images and 638 testing images.
It includes a wide variety of scenes, e.g., urban, beach, mountain, roads, parks, snow, animals, vehicles, and houses, and covers various types of pictures, e.g., aerial, landscape, close range, and selfies.
The second benchmark dataset we employed is the UCF Shadow Dataset~\cite{zhu2010learning}.
It includes 145 training images and 76 testing images, and covers outdoor scenes with various backgrounds.
We train our shadow detection network using the SBU training set.


\paragraph{Evaluation metrics.}
We employ two commonly-used metrics to quantitatively evaluate the shadow detection performance.
The first one is the accuracy metric:
\begin{equation}
accuracy \ = \ \frac{TP+TN}{N_p+N_n} \ ,
\end{equation}
where $TP$, $TN$, $N_p$ and $N_n$ are true positives, true negatives, number of shadow pixels, and number of non-shadow pixels, respectively, as defined in Section~\ref{subsec:3.2}.

Since $N_p$ is usually much smaller than $N_n$ in natural images, we employ the second metric called the balance error rate (BER) to obtain a more balanced evaluation by equally considering the shadow and non-shadow regions:
\begin{equation}
BER \ = \ (1-\frac{1}{2}(\frac{TP}{N_p}+\frac{TN}{N_n}))\times 100 \ .
\end{equation}
Note that unlike the accuracy metric, for BER, the lower its value, the better the detection result is.


\begin{table}[!t]
\begin{center}
  \caption{Comparing our method (DSC) with state-of-the-arts methods
  for shadow detection (scGAN~\cite{nguyen2017shadow}, stacked-CNN~\cite{vicente2016large}, patched-CNN~\cite{hosseinzadeh2017fast} and Unary-Pairwise~\cite{guo2011single}),
  for saliency detection (SRM~\cite{wang2017stagewise} and Amulet~\cite{zhang2017amulet}),
  and
  for semantic image segmentation (PSPNet~\cite{Zhao_2017_CVPR}).}
  \resizebox{\columnwidth}{!}{%
  \label{table:state-of-the-art}
    \begin{tabular}{c|c|c|c|c}
        &
        \multicolumn{2}{c|}{SBU~\cite{vicente2016large}} &
        \multicolumn{2}{c}{UCF~\cite{zhu2010learning}}
        \\
        \hline
        method & accuracy & BER & accuracy & BER
        \\
        \hline
        \hline
        \textbf{DSC (ours)} & \textbf{0.97} & \textbf{5.59} & \textbf{0.95} & \textbf{8.10}
        \\
        \hline
        \hline
        scGAN~\cite{nguyen2017shadow} & 0.90 & 9.10 & 0.87 & 11.50
        \\
        stacked-CNN~\cite{vicente2016large} & 0.88 & 11.00 & 0.85 & 13.00
        \\
        patched-CNN~\cite{hosseinzadeh2017fast} & 0.88 & 11.56  &-&-
        \\
        Unary-Pairwise~\cite{guo2011single} & 0.86 & 25.03 &-&-
        \\
        \hline
        \hline
        SRM~\cite{wang2017stagewise} & 0.96 & 7.25 & 0.94 & 9.81
        \\
        Amulet~\cite{zhang2017amulet} & 0.93 & 15.13 & 0.92 & 15.17
        \\
        \hline
        \hline
        PSPNet~\cite{Zhao_2017_CVPR} & 0.95 & 8.57 & 0.93 & 11.75
        \\
        \hline
    \end{tabular}
    }
  \end{center}
  \vspace{-5mm}
\end{table}


\subsection{Comparison with the State-of-the-art Shadow Detection Methods}

We compare our method with four recent shadow detection methods: scGAN~\cite{nguyen2017shadow}, stacked-CNN~\cite{vicente2016large}, patched-CNN~\cite{hosseinzadeh2017fast} and Unary-Pairwise~\cite{guo2011single}.
Among them, the first three are deep-learning-based methods, while the last one is based on hand-crafted features.
For a fair comparison, the shadow detection results of other methods are obtained either directly from results provided by the authors, or by generating them using implementations provided by the authors with recommended parameter setting.

Table~\ref{table:state-of-the-art} reports the comparison results, where we can see that our method outperforms the others in terms of both accuracy and BER for both benchmark datasets.
Note that our shadow detection network is trained using the SBU training set~\cite{vicente2016large}, but it still outperforms the others also for the UCF dataset, thus demonstrating its generalization ability.

\begin{figure*}[tp]
	\centering
	
	\vspace*{0.5mm}
	\begin{subfigure}{0.107\textwidth} 
		\includegraphics[width=\textwidth]{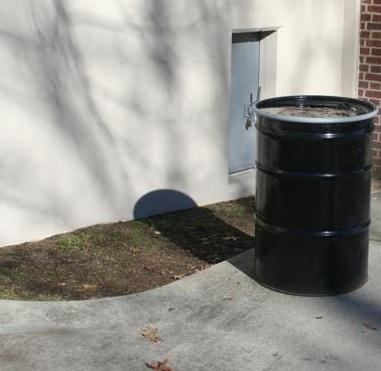}
	\end{subfigure}
	\begin{subfigure}{0.107\textwidth}
		\includegraphics[width=\textwidth]{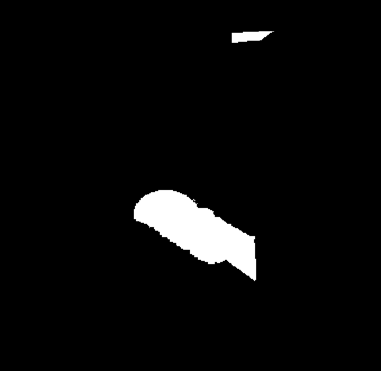}
	\end{subfigure}
	\begin{subfigure}{0.107\textwidth}
		\includegraphics[width=\textwidth]{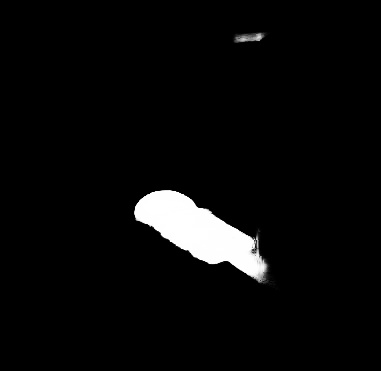}
	\end{subfigure}
	\begin{subfigure}{0.107\textwidth}
		\includegraphics[width=\textwidth]{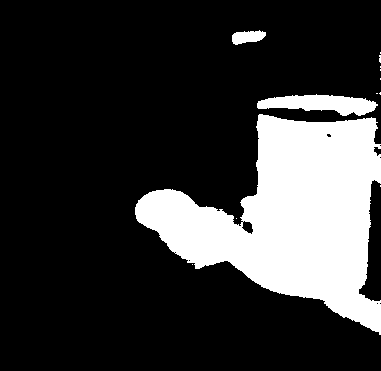}
	\end{subfigure}
	\begin{subfigure}{0.107\textwidth}
		\includegraphics[width=\textwidth]{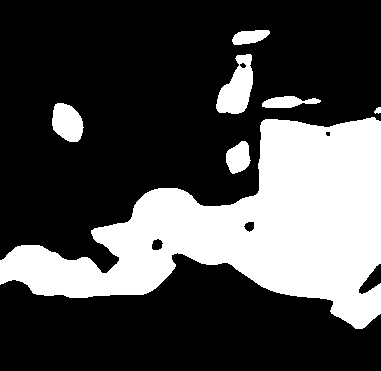}
	\end{subfigure}
	\begin{subfigure}{0.107\textwidth}
		\includegraphics[width=\textwidth]{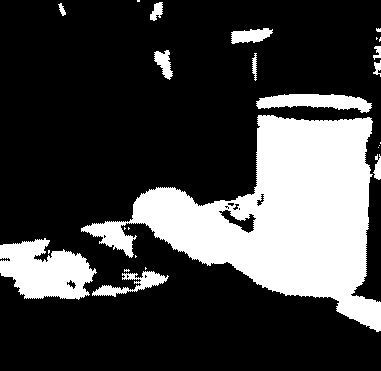}
	\end{subfigure}
	\begin{subfigure}{0.107\textwidth}
		\includegraphics[width=\textwidth]{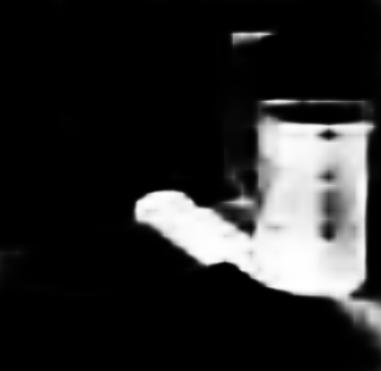}
	\end{subfigure}
	\begin{subfigure}{0.107\textwidth}
		\includegraphics[width=\textwidth]{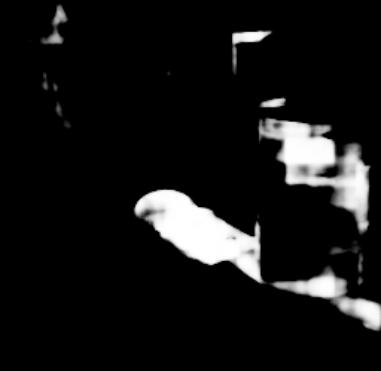}
	\end{subfigure}
	\begin{subfigure}{0.107\textwidth}
		\includegraphics[width=\textwidth]{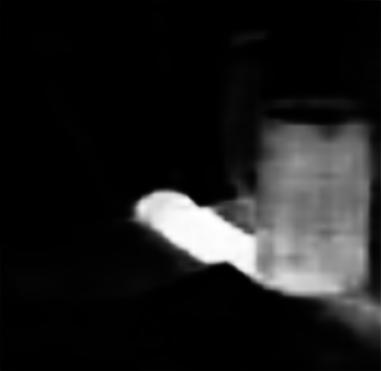}
	\end{subfigure}
	
	\ \\
	
	\vspace*{0.5mm}
	\begin{subfigure}{0.107\textwidth}
		\includegraphics[width=\textwidth]{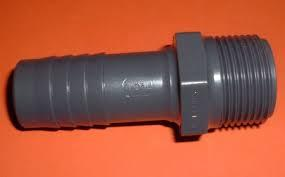}
	\end{subfigure}
	\begin{subfigure}{0.107\textwidth}
		\includegraphics[width=\textwidth]{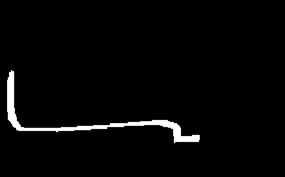}
	\end{subfigure}
	\begin{subfigure}{0.107\textwidth}
		\includegraphics[width=\textwidth]{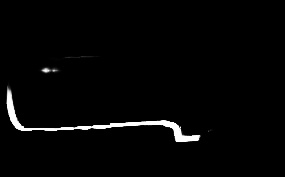}
	\end{subfigure}
	\begin{subfigure}{0.107\textwidth}
		\includegraphics[width=\textwidth]{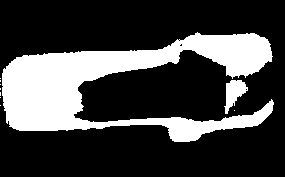}
	\end{subfigure}
	\begin{subfigure}{0.107\textwidth}
		\includegraphics[width=\textwidth]{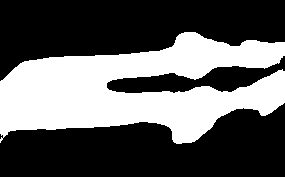}
	\end{subfigure}
	\begin{subfigure}{0.107\textwidth}
		\includegraphics[width=\textwidth]{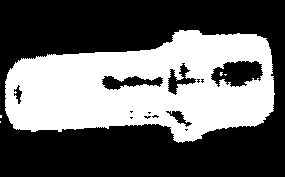}
	\end{subfigure}
	\begin{subfigure}{0.107\textwidth}
		\includegraphics[width=\textwidth]{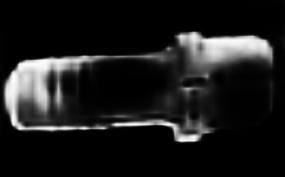}
	\end{subfigure}
	\begin{subfigure}{0.107\textwidth}
		\includegraphics[width=\textwidth]{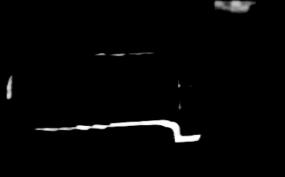}
	\end{subfigure}
	\begin{subfigure}{0.107\textwidth}
		\includegraphics[width=\textwidth]{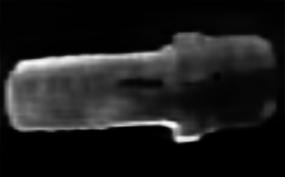}
	\end{subfigure}
	
	\ \\
	
	\vspace*{0.5mm}
	\begin{subfigure}{0.107\textwidth}
		\includegraphics[width=\textwidth]{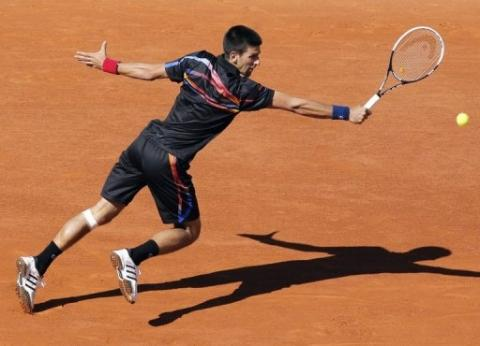}
	\end{subfigure}
	\begin{subfigure}{0.107\textwidth}
		\includegraphics[width=\textwidth]{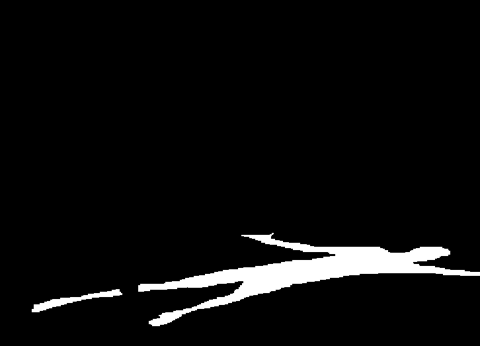}
	\end{subfigure}
	\begin{subfigure}{0.107\textwidth}
		\includegraphics[width=\textwidth]{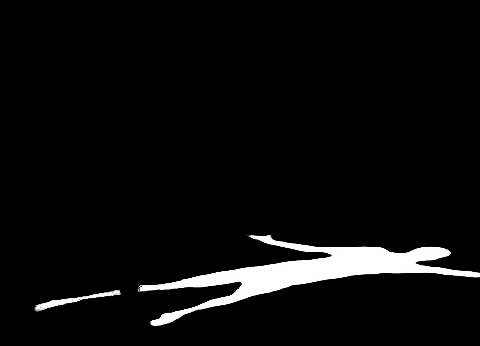}
	\end{subfigure}
	\begin{subfigure}{0.107\textwidth}
		\includegraphics[width=\textwidth]{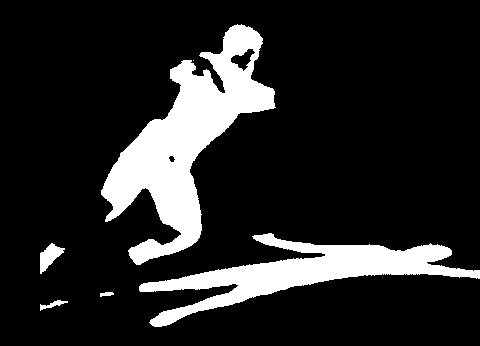}
	\end{subfigure}
	\begin{subfigure}{0.107\textwidth}
		\includegraphics[width=\textwidth]{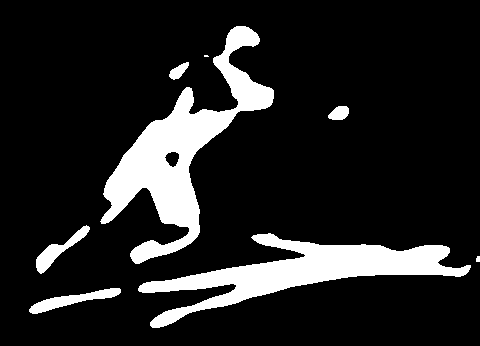}
	\end{subfigure}
	\begin{subfigure}{0.107\textwidth}
		\includegraphics[width=\textwidth]{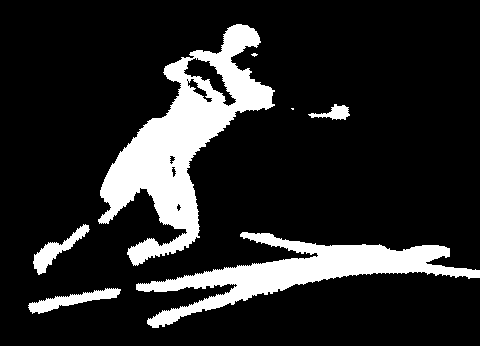}
	\end{subfigure}
	\begin{subfigure}{0.107\textwidth}
		\includegraphics[width=\textwidth]{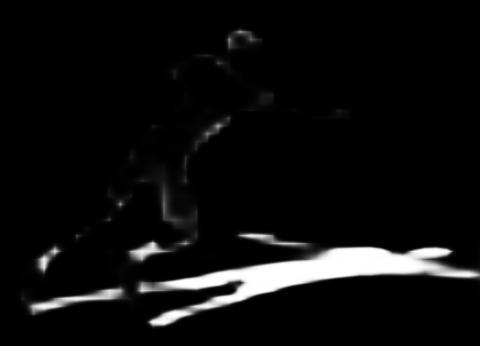}
	\end{subfigure}
	\begin{subfigure}{0.107\textwidth}
		\includegraphics[width=\textwidth]{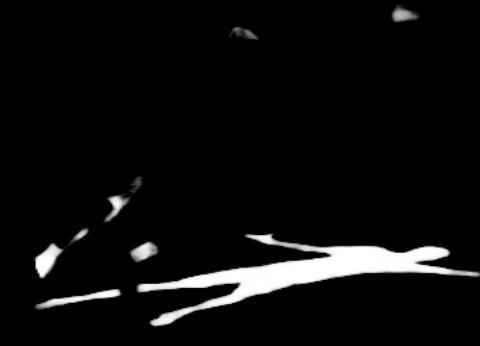}
	\end{subfigure}
	\begin{subfigure}{0.107\textwidth}
		\includegraphics[width=\textwidth]{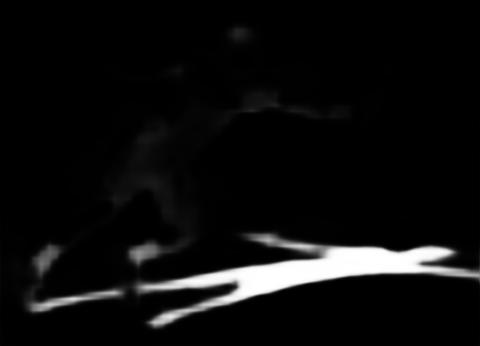}
	\end{subfigure}
	
	\ \\
	\vspace*{0.5mm}
	\begin{subfigure}{0.107\textwidth}
		\includegraphics[width=\textwidth]{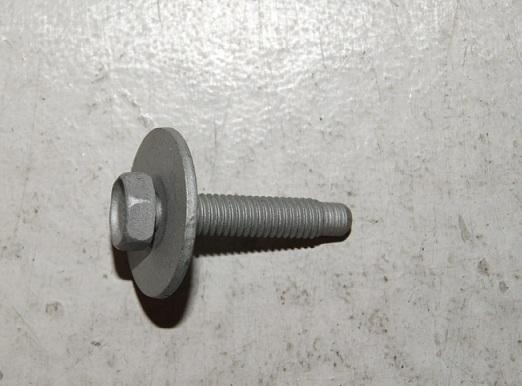}
	\end{subfigure}
	\begin{subfigure}{0.107\textwidth}
		\includegraphics[width=\textwidth]{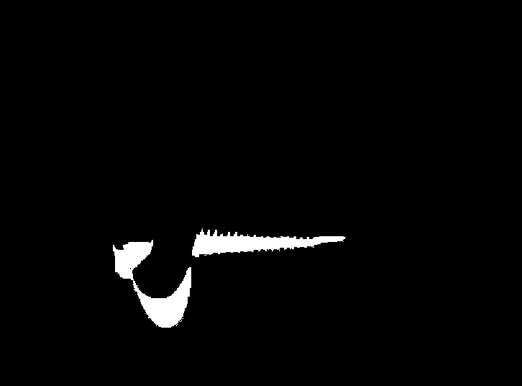}
	\end{subfigure}
	\begin{subfigure}{0.107\textwidth}
		\includegraphics[width=\textwidth]{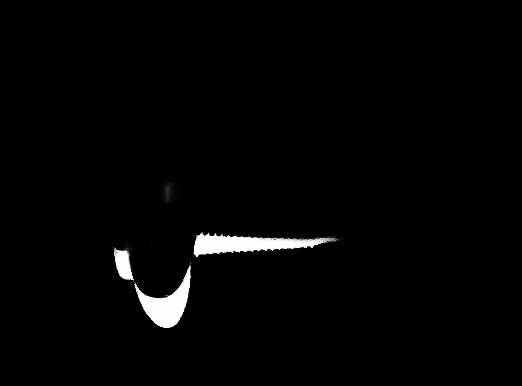}
	\end{subfigure}
	\begin{subfigure}{0.107\textwidth}
		\includegraphics[width=\textwidth]{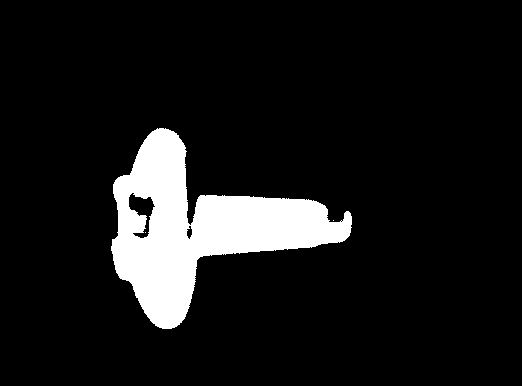}
	\end{subfigure}
	\begin{subfigure}{0.107\textwidth}
		\includegraphics[width=\textwidth]{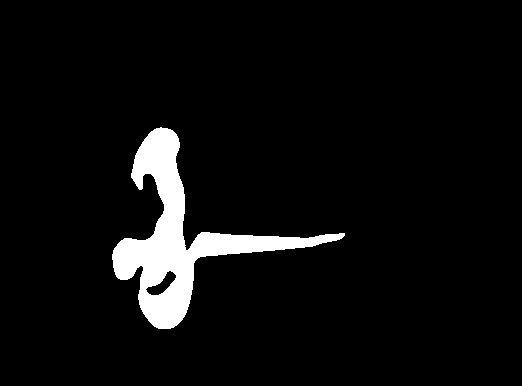}
	\end{subfigure}
	\begin{subfigure}{0.107\textwidth}
		\includegraphics[width=\textwidth]{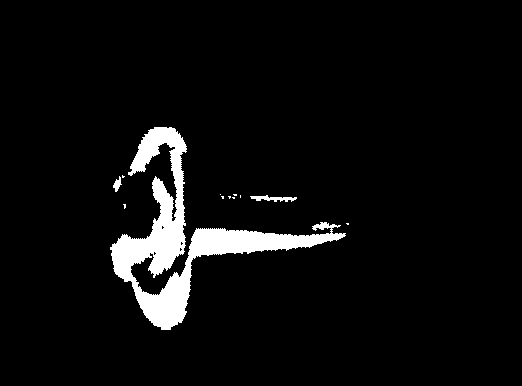}
	\end{subfigure}
	\begin{subfigure}{0.107\textwidth}
		\includegraphics[width=\textwidth]{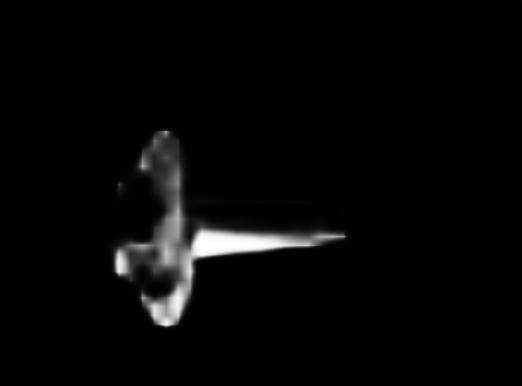}
	\end{subfigure}
	\begin{subfigure}{0.107\textwidth}
		\includegraphics[width=\textwidth]{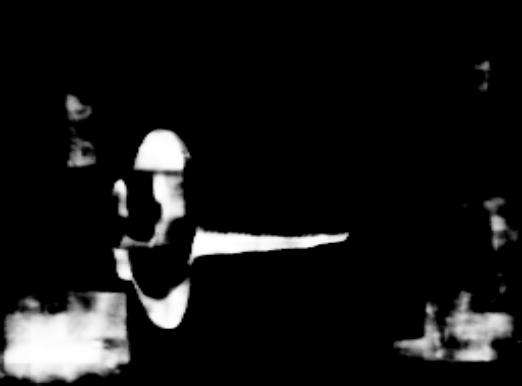}
	\end{subfigure}
	\begin{subfigure}{0.107\textwidth}
		\includegraphics[width=\textwidth]{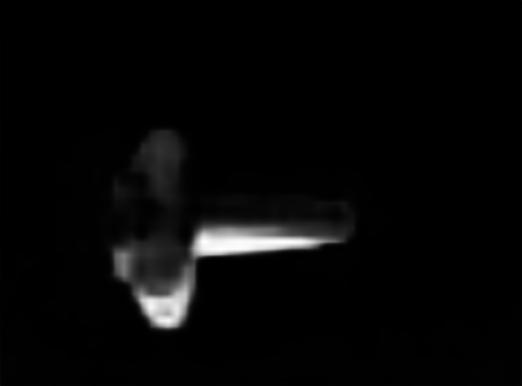}
	\end{subfigure}

	\vspace*{0.5mm}
	\begin{subfigure}{0.107\textwidth}
		\includegraphics[width=\textwidth]{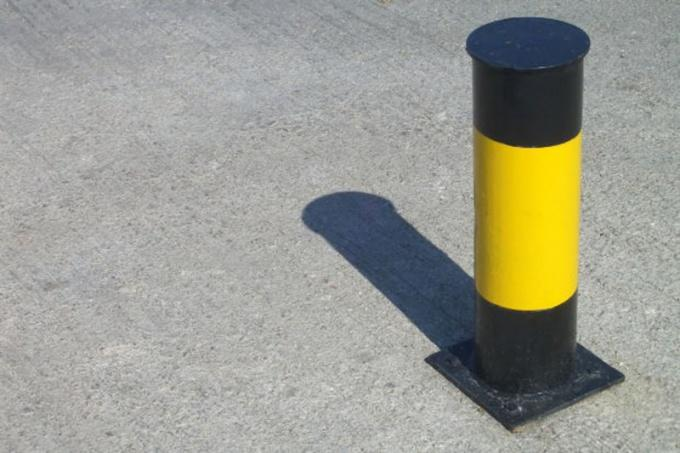}
		\vspace{-5.5mm} \caption*{{\footnotesize input image}}
	\end{subfigure}
	\begin{subfigure}{0.107\textwidth}
		\includegraphics[width=\textwidth]{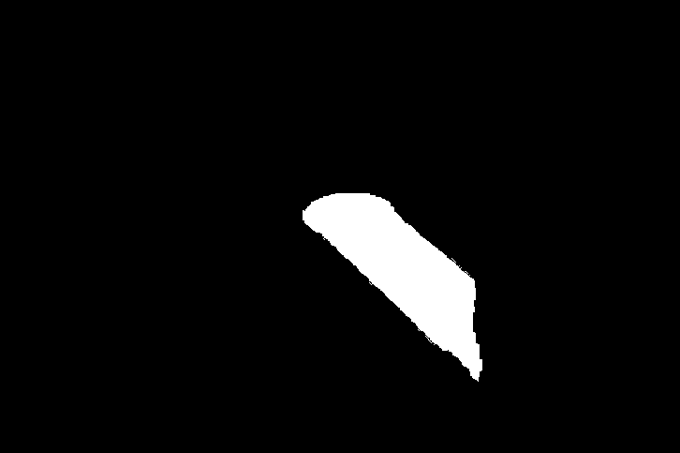}
		\vspace{-5.5mm} \caption*{{\footnotesize ground truth}} 
	\end{subfigure}
	\begin{subfigure}{0.107\textwidth}
		\includegraphics[width=\textwidth]{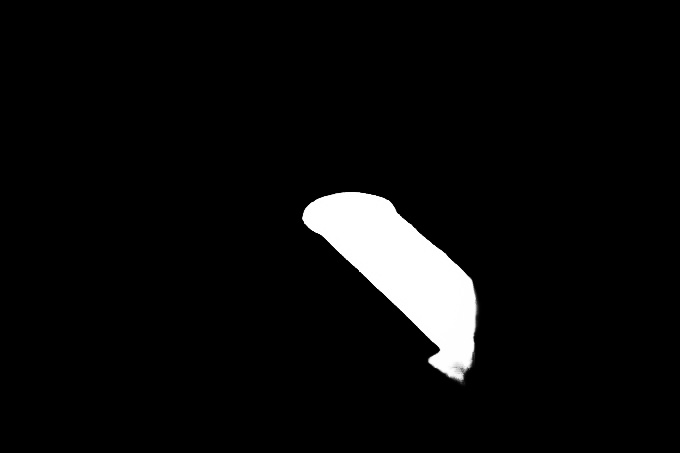}
		\vspace{-5.5mm} \caption*{{\footnotesize DSC (ours)}} 
	\end{subfigure}
	\begin{subfigure}{0.107\textwidth}
		\includegraphics[width=\textwidth]{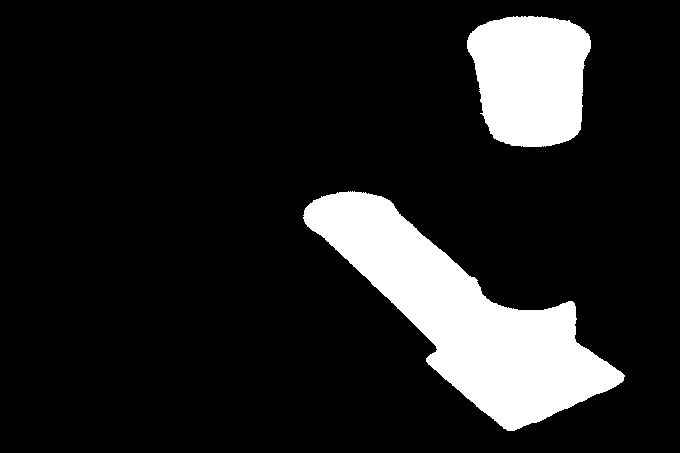}
		\vspace{-5.5mm} \caption*{{\footnotesize scGAN~\cite{nguyen2017shadow}}}
	\end{subfigure}
	\begin{subfigure}{0.107\textwidth}
		\includegraphics[width=\textwidth]{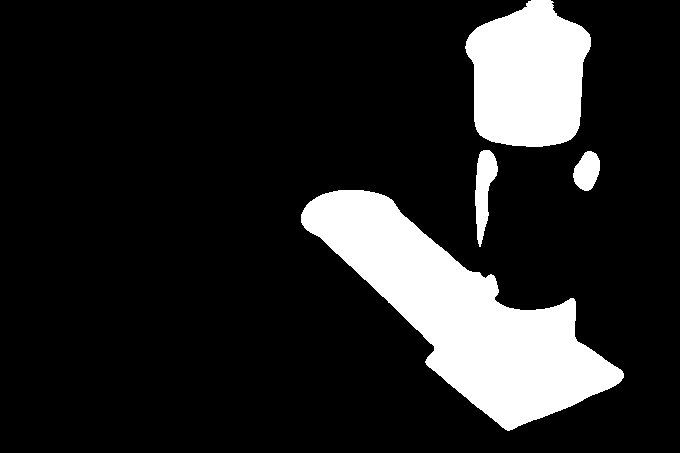}
		\vspace{-5.5mm} \caption*{{\footnotesize stkd'-CNN~\cite{vicente2016large}}}
	\end{subfigure}
	\begin{subfigure}{0.107\textwidth}
		\includegraphics[width=\textwidth]{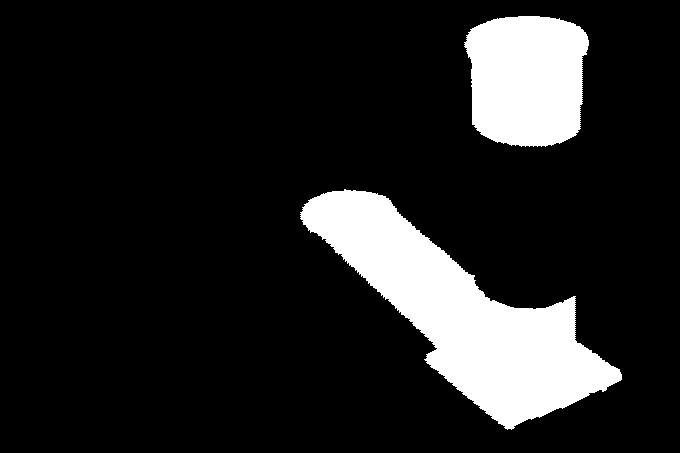}
		\vspace{-5.5mm} \caption*{{\footnotesize patd'-CNN~\cite{hosseinzadeh2017fast}}}
	\end{subfigure}
	\begin{subfigure}{0.107\textwidth}
		\includegraphics[width=\textwidth]{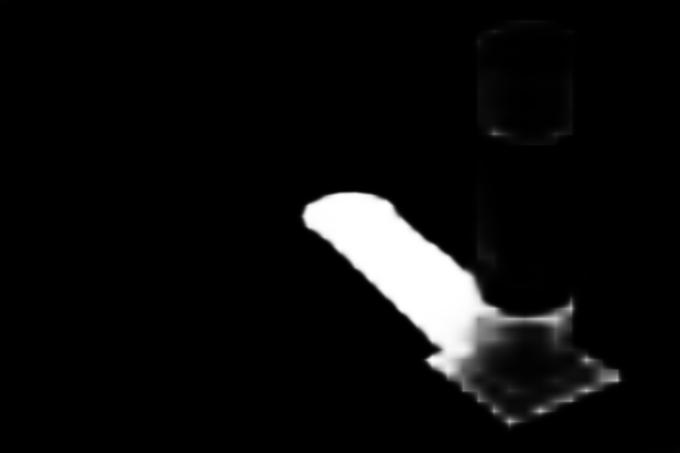}
		\vspace{-5.5mm} \caption*{{\footnotesize SRM~\cite{wang2017stagewise}}}
	\end{subfigure}
	\begin{subfigure}{0.107\textwidth}
		\includegraphics[width=\textwidth]{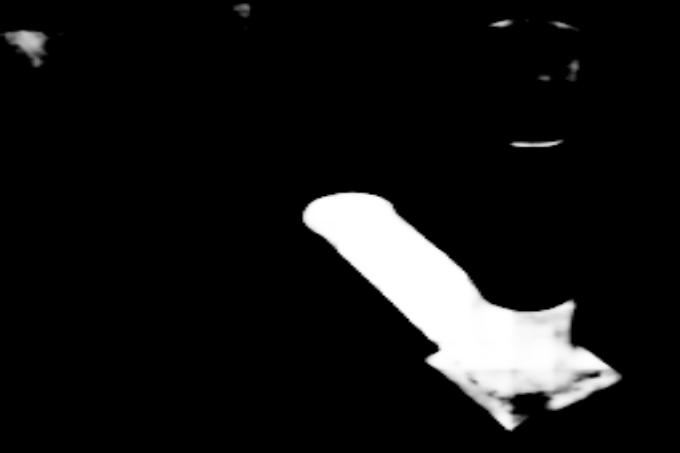}
		\vspace{-5.5mm} \caption*{{\footnotesize Amulet~\cite{zhang2017amulet}}}
	\end{subfigure}
	\begin{subfigure}{0.107\textwidth}
		\includegraphics[width=\textwidth]{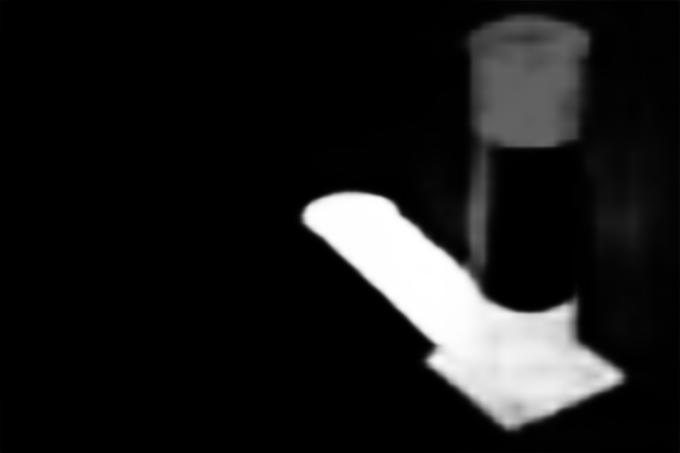}
		\vspace{-5.5mm} \caption*{\hspace*{-0.7mm}{\footnotesize PSPNet~\cite{Zhao_2017_CVPR}}}
	\end{subfigure}

	\vspace{-1.0mm}
	\caption{More visual comparison results (continue from Figure~\ref{fig:comparison_real_photos}).}
	\label{fig:comparison_real_photos2}
	\vspace{-2.5mm}
\end{figure*}

We further provide visual comparison results in Figures~\ref{fig:comparison_real_photos} and~\ref{fig:comparison_real_photos2}, which show various challenging cases, e.g., a light shadow next to a dark shadow, shadows around complex backgrounds, and black objects around shadows.
Without understanding the global image semantics, it is hard to locate these shadows, and non-shadow regions would be easily misrecognized as shadows.
From the results, we can see that our method can effectively locate shadows and avoid false positives as compared to the others; for black objects misrecognized as shadows by other methods, our method could still recognize them as non-shadows.

\subsection{Comparison with Saliency Detection and Semantic Segmentation Methods}
In general, deep networks designed for saliency detection and semantic image segmentation may also be used for shadow detection by training the networks using datasets of annotated shadows.
Hence, we conduct another experiment by using two recent deep models for saliency detection, i.e., SRM~\cite{wang2017stagewise} and Amulet~\cite{zhang2017amulet}, and a recent deep model for semantic image segmentation, i.e., PSPNet~\cite{Zhao_2017_CVPR}.

For a fair comparison, we re-train their models on the SBU training set using implementations provided by the authors, and adjust the training parameters to obtain the best shadow detection results.
The last three rows in Table~\ref{table:state-of-the-art} report the comparison results in terms of the accuracy and BER metrics.
Although these methods achieve good results for both metrics, our method still outperforms them for both benchmark datasets.
Please also refer to the last three columns in Figures~\ref{fig:comparison_real_photos} and~\ref{fig:comparison_real_photos2} for visual comparison results.


\subsection{Evaluation on the DCS Module}


\begin{table}  [tp]
	\begin{center}
		\caption{Component analysis.
			We train three networks using the SBU training set and test them using the SBU testing set~\cite{vicente2016large}:
			``basic'' denotes the architecture shown in Figure~\ref{fig:context} but without all DSC modules;
			``basic+context'' denotes the ``basic'' network with spatial context but not direction-aware spatial context; and
			``DSC" is the overall network in Figure~\ref{fig:context}.}
		\label{table:ablation}
		\begin{tabular}{c|c|c}
			\hline
			network & BER & improvement \\ 
			\hline
			basic &  6.55 & -\\ 
			
			basic+context & 6.23 & 4.89\% \\ 
			
			DSC & \textbf{5.59} & 10.27\% \\ 
			
			\hline
			
		\end{tabular} 
	\end{center}
\vspace{-2mm}
\end{table}

\begin{figure} [tp]
	\centering
	\includegraphics[width=0.98\linewidth]{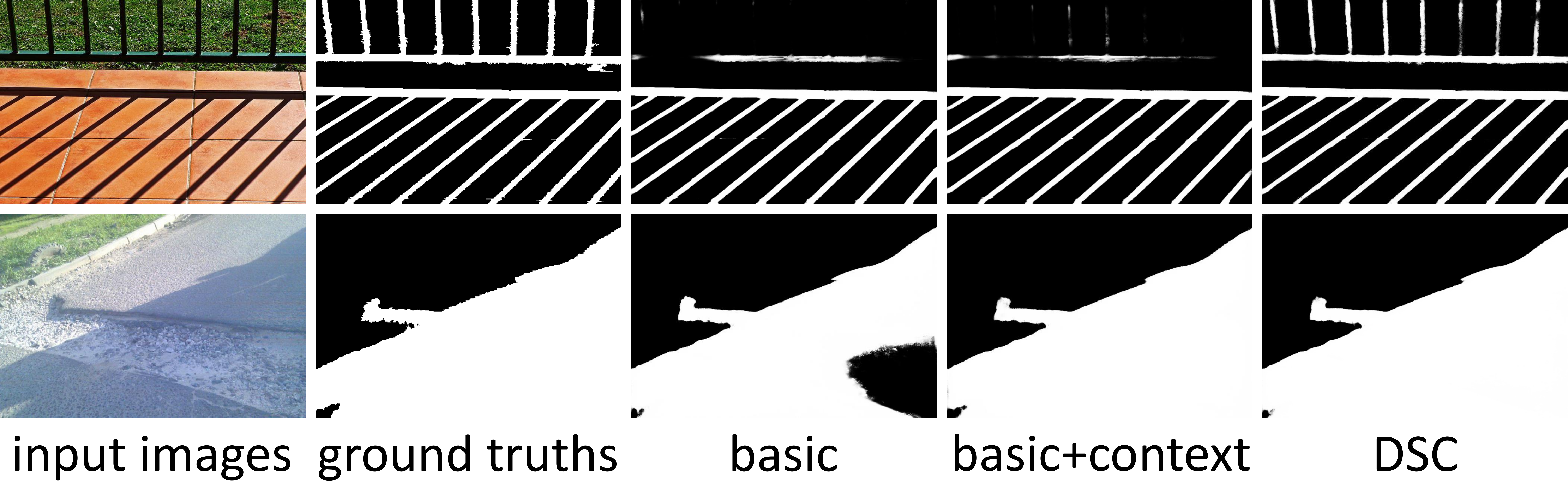}
	\caption{Visual comparison results of component analysis.}
	\label{fig:component_analysis}
	\vspace{-2mm}
\end{figure}

\begin{table}  [tp]
	\begin{center}
		\caption{DSC architecture analysis.
			By varying the parameters in the DSC architecture (see 2nd and 3rd columns below), we can have produce a slightly different overall network and explore their performance (see last column).}
		\label{table:rnn}
		\begin{tabular}{c|c|c}
			\hline
			number of rounds & shared $\mathbf{W}? $ & BER  \\
			\hline
			1 & - & 5.85
			\\
			2 & Yes & \textbf{5.59}
			\\
			3 & Yes & 5.85
			\\
			\hline
			2 & No & 6.02
			\\
			\hline			
		\end{tabular} 
	\end{center}
\vspace{-3mm}
\end{table}

\paragraph{Component analysis.}
We perform an experiment to evaluate the effectiveness of the DSC module design.
Here, we use the SBU dataset and consider two baseline networks.
The first baseline (denoted as ``basic'') is a network constructed by removing all the DSC modules from the overall network shown in Figure~\ref{fig:arc}.
The second baseline (denoted as ``basic+context") considers spatial context but ignores the direction-aware attention weights.
Compared with the first baseline, this network has all the DSC modules, but it removes the direction-aware attention mechanism inside the DSC modules, i.e., removing the computation of $\mathbf{W}$ and directly concatenating the context features without multiplying them with the attention weights; see Figure~\ref{fig:context}.
This is equivalent to setting all the attention weights $\mathbf{W}$ to be one.

Table~\ref{table:ablation} reports the comparison results, showing that our basic network with multi-scale features and the weighed cross entropy loss function can produce good results.
Moreover, by considering spatial context and DSC features can lead to further obvious improvement.
See also Figure~\ref{fig:component_analysis} for visual comparison results.


\paragraph{DSC architecture analysis.}
When we design the network structure in the DSC module, we encounter two questions:
(i) how many rounds of recurrent translations we should employ in the spatial RNN; and 
(ii) whether to share the attention weights, or to use separate attention weights in different rounds of recurrent translations in the spatial RNN.

We modify our network for these two parameters and produce the quantitative comparison results shown in Table~\ref{table:rnn}.
From the results, we can see that having two rounds of recurrent translations and sharing the attention weights in both rounds produce the best detection result.
We believe that when there is only one round of recurrent translations, the global context information cannot well propagate over the spatial domain, so there is insufficient information exchange for learning the shadows.
On the other hand, when we have three rounds of recurrent translations or separate copies of attention weights, we will end up having too many parameters in the network, making it hard to be trained.

\begin{figure} [tp]
	\centering
	\includegraphics[width=0.98\linewidth]{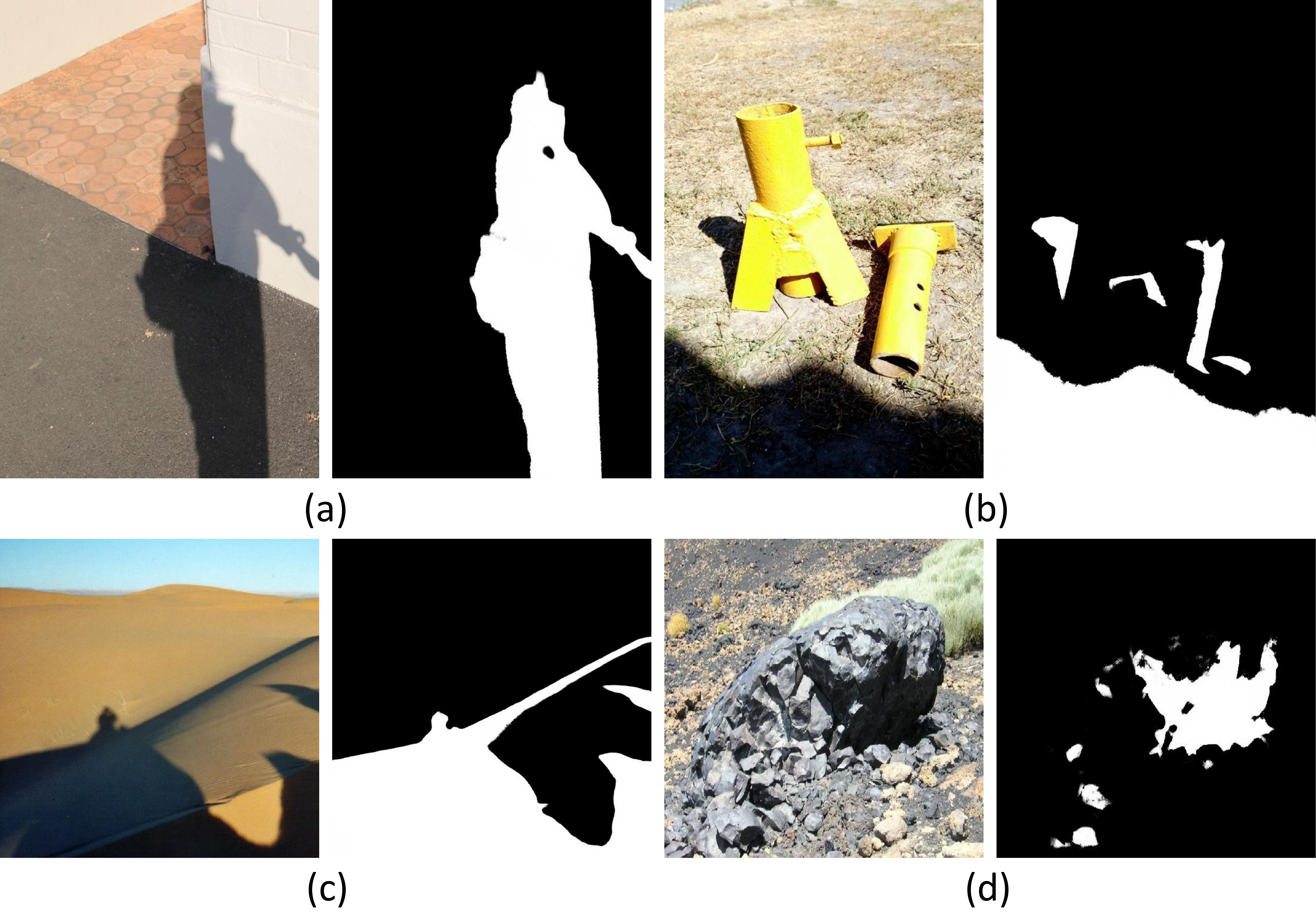}
	\caption{More results produced from our method.}
	\label{fig:more_results}
\end{figure}

\begin{figure} [tp]
	\centering
	\includegraphics[width=0.98\linewidth]{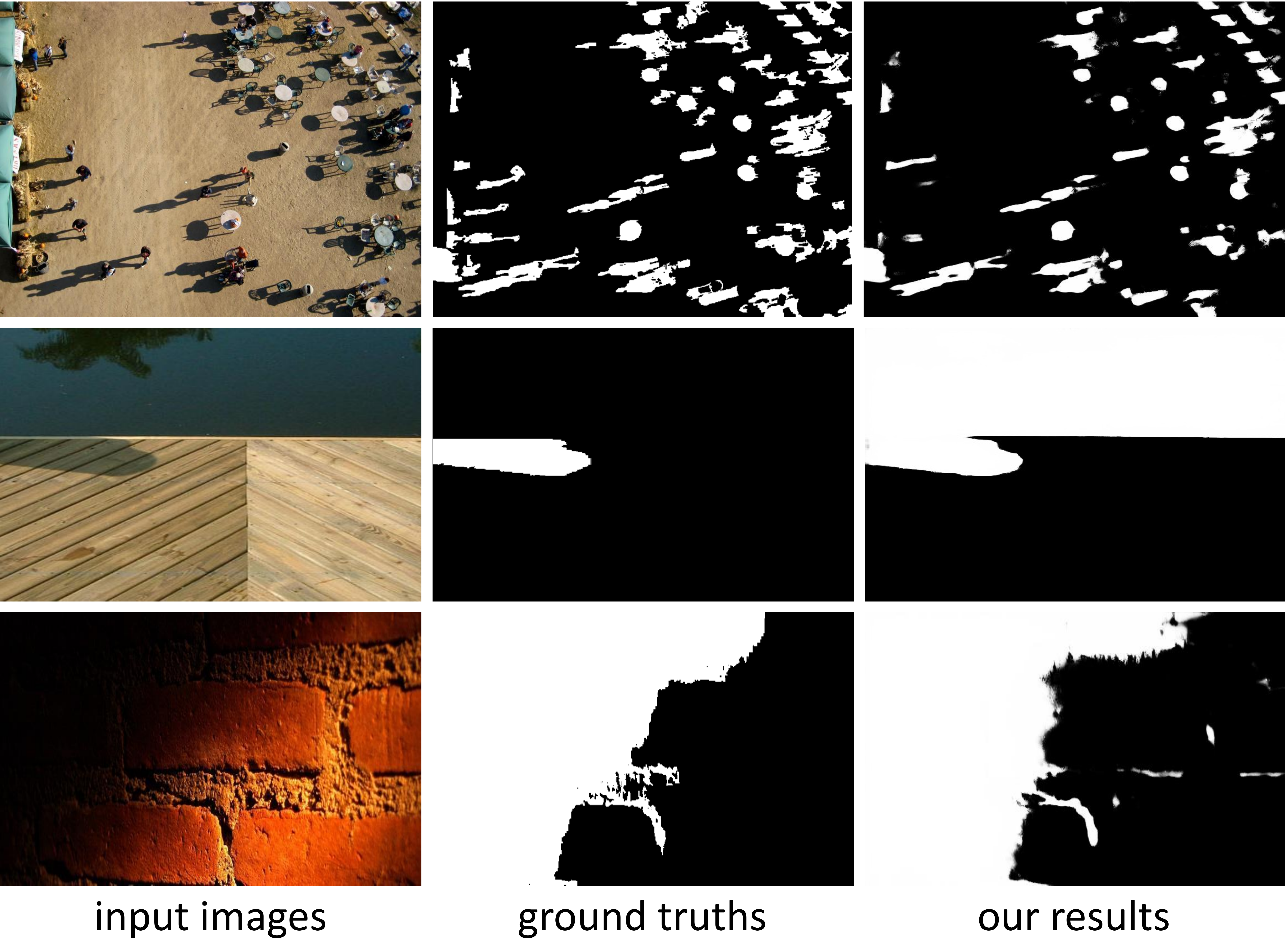}
	\caption{Failure cases.}
	\label{fig:fail_case}
\end{figure}

\subsection{More Shadow Detection Results}

Figure~\ref{fig:more_results} shows more shadow detection results:
(a) light and dark shadows locate next to each other;
(b) small and unconnected shadows;
(c) no clear boundary between shadow and non-shadow regions; and
(d) shadows of irregular shapes.
Our method can still detect these shadows fairly well, but it fails in some extremely complex scenes:
(a) a scene with many small shadows (see $1^{st}$ row in Figure~\ref{fig:fail_case}), where the features in deep layers lose the detail information and features in shallow layers lack the semantic information for the shadow context;
(b) a scene with a large black region (see $2^{nd}$ row in Figure~\ref{fig:fail_case}), where there are insufficient surrounding context to indicate whether it is a shadow or simply a black object; and
(c) a scene with soft shadows (see $3^{rd}$ row in Figure~\ref{fig:fail_case}), where the difference between the soft shadow regions and the non-shadow regions is small.
The code, trained model, and more shadow detection results on the datasets are publicly available at \textit{\url{https://xw-hu.github.io/}}.

\section{Conclusion}
\label{sec::conclusion}

This paper presents a novel network for single-image shadow detection by harvesting direction-aware spatial context.
Our key idea is to analyze multi-level spatial context in a direction-aware manner by formulating a direction-aware attention mechanism in a spatial RNN.
In our mechanism, the network can automatically learn the attention weights for leveraging and composing the spatial context in different directions in the spatial RNN.
In this way, we can produce direction-aware spatial context (DSC) features and formulate the DSC module for the task.
Further, we adopt multiple DSC modules in a multi-layer convolutional neural network to predict score maps in different scales, and design a weighted cross entropy loss function to make effective the training process.
In the end, we test our network on two benchmark datasets, compare it with various state-of-the-art methods, and show the superiority of our network over the others in terms of the accuracy and BER metrics.

In future, we plan to explore the potential of our network for other applications such as saliency detection and semantic segmentation, and further enhance its capability for detecting time-varying shadows in videos.

\section*{Acknowledgments}
The work is supported by the National Basic Program of China, 973 Program (Project no. 2015CB351706), the Research Grants Council of the Hong Kong Special Administrative Region (Project no. CUHK 14225616), the Shenzhen Science and Technology Program (No. JCYJ20170413162617606), the CUHK strategic recruitment fund, and the Innovation and Technology Fund of Hong Kong (Project no. ITS/304/16).
We thank reviewers for their valuable comments, Michael S. Brown for his discussion, and Minh Hoai Nguyen for sharing their results.
Xiaowei Hu is funded by the Hong Kong Ph.D. Fellowship.

{\small
\bibliographystyle{ieee}
\bibliography{egbib}
}

\end{document}